\documentclass[sigconf]{acmart}

\usepackage{multirow}
\usepackage{float}
\usepackage{hyperref}
\hypersetup{
    colorlinks=true,
    urlcolor=blue,
    linkcolor=red,
    citecolor=green
}
\usepackage{subcaption}

\AtBeginDocument{%
  }

\copyrightyear{2026}
\acmYear{2026}
\setcopyright{cc}
\setcctype{by}
\acmConference[CHI '26]{Proceedings of the 2026 CHI Conference on Human Factors in Computing Systems}{April 13--17, 2026}{Barcelona, Spain}
\acmBooktitle{Proceedings of the 2026 CHI Conference on Human Factors in Computing Systems (CHI '26), April 13--17, 2026, Barcelona, Spain}
\acmPrice{}
\acmDOI{10.1145/3772318.3791338}
\acmISBN{979-8-4007-2278-3/2026/04}

\begin{document}

\title{Game-Based and Gamified Robotics Education: A Comparative Systematic Review and Design Guidelines}


\author{Syed Tanzim Mubarrat}
\email{smubarra@purdue.edu}
\orcid{0000-0002-9702-8901}
\affiliation{%
  \institution{Purdue University}
  \streetaddress{610 Purdue Mall}
  \city{West Lafayette}
  \state{Indiana}
  \country{USA}
  \postcode{47907}
}

\author{Byung-Cheol Min}
\email{minb@iu.edu}
\orcid{0000-0001-6458-4365}
\affiliation{%
  \institution{Indiana University Bloomington}
  \streetaddress{107 S Indiana Ave}
  \city{Bloomington}
  \state{Indiana}
  \country{USA}
  \postcode{47405}
}

\author{Tianyu Shao}
\email{shao122@purdue.edu}
\orcid{0009-0006-7236-0357}
\affiliation{%
  \institution{Purdue University}
  \streetaddress{610 Purdue Mall}
  \city{West Lafayette}
  \state{Indiana}
  \country{USA}
  \postcode{47907}
}

\author{E. Cho Smith}
\email{wilso287@purdue.edu}
\orcid{0009-0008-7817-690X}
\affiliation{%
  \institution{Purdue University}
  \streetaddress{610 Purdue Mall}
  \city{West Lafayette}
  \state{Indiana}
  \country{USA}
  \postcode{47907}
}

\author{Bedrich Benes}
\email{bbenes@purdue.edu}
\orcid{0000-0002-5293-2112}
\affiliation{%
  \institution{Purdue University}
  \streetaddress{610 Purdue Mall}
  \city{West Lafayette}
  \state{Indiana}
  \country{USA}
  \postcode{47907}
}

\author{Alejandra J. Magana}
\email{admagana@purdue.edu}
\orcid{0000-0001-6117-7502}
\affiliation{%
  \institution{Purdue University}
  \streetaddress{610 Purdue Mall}
  \city{West Lafayette}
  \state{Indiana}
  \country{USA}
  \postcode{47907}
}

\author{Christos Mousas}
\email{cmousas@purdue.edu}
\orcid{0000-0003-0955-7959}
\affiliation{%
  \institution{Purdue University}
  \streetaddress{610 Purdue Mall}
  \city{West Lafayette}
  \state{Indiana}
  \country{USA}
  \postcode{47907}
}

\author{Dominic Kao}
\email{kalikao@illinois.edu}
\orcid{0000-0002-7732-6258}
\affiliation{%
  \institution{\makebox[0pt][c]{University of Illinois Urbana-Champaign}}
  \streetaddress{201 N Goodwin Ave}
  \city{Urbana}
  \state{Illinois}
  \country{USA}
  \postcode{61801}
}

\renewcommand{\shortauthors}{Mubarrat et al.}

\begin{abstract}

  Robotics education fosters computational thinking, creativity, and problem-solving, but remains challenging due to technical complexity. Game-based learning (GBL) and gamification offer engagement benefits, yet their comparative impact remains unclear. We present the first PRISMA-aligned systematic review and comparative synthesis of GBL and gamification in robotics education, analyzing 95 studies from 12{,}485 records across four databases (2014--2025). We coded each study’s approach, learning context, skill level, modality, pedagogy, and outcomes ($\kappa = .918$). Three patterns emerged: (1) approach--context--pedagogy coupling (GBL more prevalent in informal settings, while gamification dominated formal classrooms [$p < .001$] and favored project-based learning [$p = .009$]); (2) emphasis on introductory programming and modular kits, with limited adoption of advanced software (\textasciitilde17\%), advanced hardware (\textasciitilde5\%), or immersive technologies (\textasciitilde22\%); and (3) short study horizons, relying on self-report. We propose eight research directions and a design space outlining best practices and pitfalls, offering actionable guidance for robotics education.

\end{abstract}



\maketitle

\section{INTRODUCTION}
In recent years, the increased focus on STEM (Science, Technology, Engineering, Mathematics) education has ignited greater interest in robotics as a viable solution to real-world challenges~\cite{rockland_advancing_2010,zeidler_stem_2016}. Robotics education is increasingly recognized as a powerful means of fostering 21st-century skills, such as computational thinking, problem-solving, systems thinking, collaboration, and creativity~\cite{alimisis_educational_2013,bers_computational_2014}. As robotics inherently integrates multiple domains (including programming, electronics, and embedded systems), it offers a multidisciplinary platform for experiential learning. However, for many learners, particularly novices or those in non-traditional settings, robotics can appear intimidating or inaccessible due to the technical complexity and steep learning curve involved~\cite{eguchi_robotics_2014,nugent_impact_2012,watanabe_compulsory_2018}. Consequently, cultivating individuals' interest in learning robotics proves to be challenging. Motivation emerges as a pivotal factor in this scenario, given its significant role in enhancing cognitive strategies~\cite{linnenbrink_motivation_2002,lynch_motivational_2006,mouton_synergogy_1984,ryan_self-determination_2000}.

Game-based learning (GBL) and gamification have emerged as promising pedagogical strategies to make STEM education more engaging, motivating, and accessible~\cite{andrade2020evaluating,sailer_gamification_2020}, notably in programming~\cite{costa_using_2023,de_paula_playing_2018,kao_hack_2020,lee_principles_2014,lee_comparing_2015,smiderle_impact_2020} and electronics~\cite{dewantara_game-based_2021,luthon_laboremremote_2014,salas-rueda_construction_2022}. GBL involves the use of fully developed games for learning or teaching applications~\cite{deterding_game_2011,sailer_gamification_2020}, whereas gamification refers to the integration of game-like elements, such as points, badges, leaderboards, or challenges, into non-game learning environments to create a modified version perceived by users as resembling a game~\cite{bahrin_enjoyment_2021,sailer_gamification_2020}. These approaches have shown potential in improving not only learner engagement~\cite{andrade2020evaluating} and motivation~\cite{stiller_game-based_2019,friehs2020effective} but also knowledge retention and skill acquisition~\cite{clark_digital_2016,papastergiou_digital_2009,khade2022requirements}.

Although GBL and gamified approaches are increasingly recognized for boosting engagement, collaboration, and creativity in robotics education, existing evidence remains fragmented. Quantitative and thematic reviews show that both GBL and gamification in robotics education can increase motivation and creativity~\cite{adipat_engaging_2021,chen_gamified_2023} and that game structures such as quest-based, competitive, and narrative modes foster both emotional engagement and practical skill gains~\cite{fante_navigating_2024,kasenides_edifying_2024}. Broader reviews of programming and computer science GBL echo these benefits but criticize the predominance of one-off studies with minimal robotics integration~\cite{miljanovic_review_2018,videnovik_game-based_2023}. A recent multilevel meta-analysis confirms moderate overall effects of educational robotics on STEM learning, but also flags wide design heterogeneity~\cite{ouyang_effects_2024}. Together, these syntheses reveal four persistent gaps: limited use of immersive technologies (such as virtual reality [VR] and haptics); scarce exploration of informal learning settings; few designs that teach programming, software development, and understanding of robotic systems (control, sensors, and integration) in a single framework; and no systematic comparison of GBL versus gamified approaches in robotics education.


Addressing these gaps requires disentangling how game mechanics, robot types, learning environments, and multimodal interfaces jointly shape outcomes. A recent study by Heinze et al.~\cite{alves2024exploring} showed that action-intensive mechanics boosted mastery and perceived control but offered less relaxation, showing that specific mechanics, not just games in general, determine motivational and affective outcomes. Moreover, immersive interfaces are highlighted as an untapped opportunity for deeper embodiment and implicit skill transfer, while informal contexts could widen participation beyond formal classrooms. A comprehensive review that systematically codes these variables will therefore advance both theory and practice, guiding the design of more inclusive, engaging, and sustainable robotics-learning experiences. 


Through this systematic literature review, we investigate the following research questions:
\begin{itemize}
\item {\textbf{RQ1}}: What game-based and gamified approaches have been employed to teach robotics and associated skills (i.e., programming, software development, and understanding of robotic systems), and what trends can be observed across the core study features?\footnote{We use the term \textit{core study features} to refer to: the type of approach (game-based or gamified), the learning context (formal or informal), applied pedagogical models, programming/software development skill levels addressed, understanding of robotic systems skills addressed, intervention modalities, usage of immersive technology, the type and experience level of the intended audience,  specific gamification elements employed, and game genres used. We selected this broad set of features based on existing literature and expert consultation with researchers in HCI, robotics, and games, as these features are necessary to provide a rigorous and holistic summary of the field. We define them in detail in Section~\ref{background}, in their respective subsections in Section~\ref{results_studyChars}, and in the \href{https://osf.io/gq3a8/files/pnfy2}{Supplementary Materials}.}
\item {\textbf{RQ2}}: How do the different types of core study features differ in their effects on learning outcomes, motivation, and other affective outcomes?
\item {\textbf{RQ3}}: What are the limitations and challenges of game-based and gamified approaches to teaching robotics and associated skills? 
\end{itemize}


Our Preferred Reporting Items for Systematic Reviews and Meta-Analyses (PRISMA)-aligned systematic review is designed to bridge these gaps by coding technology stacks (e.g., robot platform, programming interface, electronic components), recording multimodal and immersive features (e.g., VR, haptics), distinguishing formal versus informal venues, capturing detailed game mechanics and narrative structures, and assessing differential outcomes for GBL and gamification across multiple robotics related skills.

Our contribution here offers the first comparative synthesis of GBL and gamification in robotics education. We (1) systematically searched four databases (2014--2025), screened 12,485 records, and included 95 studies; (2) coded approach, learning context, skill level, intervention modality, pedagogy, and outcomes with very high inter-rater reliability ($\kappa = .918$); (3) observed robust patterns--approach--context coupling, over-emphasis on introductory programming and modular kits with limited advanced software (\textasciitilde17\%), advanced hardware (\textasciitilde5\%), and immersive technologies (\textasciitilde22\%), and short study horizons with reliance on self-report; and (4) synthesized a design space with recommended practices and pitfalls and identify eight concrete future research directions. The result is actionable guidance for educators and designers while sharpening theoretical accounts of how play, embodiment, and context co-produce robotics education.




\section{BACKGROUND AND RELATED WORK}\label{background}



\subsection{Game-based Learning and Gamification}\label{background_approach}

GBL and gamification have emerged as influential educational approaches designed to enhance engagement, motivation, and accessibility within STEM education~\cite{deterding_game_2011,sailer_gamification_2020}. Specifically, these methodologies have seen notable applications in teaching programming concepts~\cite{costa_using_2023,de_paula_playing_2018,kao_hack_2020,lee_principles_2014,lee_comparing_2015,smiderle_impact_2020}, with substantial evidence indicating their effectiveness in improving learner outcomes. For instance, Costa~\cite{costa_using_2023} conducted a multi-level meta-analysis demonstrating that integrating game concepts into programming education significantly increased students' understanding and retention.


In the domain of electronics education, GBL and gamified approaches have shown similar potential for elevating student performance and motivation~\cite{dewantara_game-based_2021,luthon_laboremremote_2014,salas-rueda_construction_2022}. Dewantara et al.~\cite{dewantara_game-based_2021} found that game-based learning significantly improved higher-order thinking and problem-solving skills in prospective physics teachers during a digital electronics course. Similarly, Luthon and Larroque~\cite{luthon_laboremremote_2014} developed LaboREM, a remote electronics lab with game-like features, which boosted engagement, interactivity, and practical skills compared to traditional methods.


Beyond cognitive benefits, gamified learning environments have been recognized for their positive impact on learners' motivational and behavioral dimensions. Smiderle et al.~\cite{smiderle_impact_2020} showed that tailoring gamification to personality traits improved engagement and adaptive behaviors. Sailer and Homner’s meta-analysis confirmed gamification boosts motivation and long-term retention~\cite{sailer_gamification_2020}. Similarly, Andrade et al.~\cite{andrade2020evaluating} found that gamification elements such as points, badges, and leaderboards positively affect engagement, usability, and STEM learning performance.




\subsection{Learning Context: Formal and Informal}\label{background_context}

Formal learning occurs in structured settings such as schools or training programs, where qualified instructors follow approved curricula and assess progress against set standards~\cite{johnson_formal_2022}. Such environments often require compulsory or credit-bearing participation and offer what Golding et al.~\cite{golding_informal_2009} term a tightly bounded ``institutional frame.'' Informal learning, by contrast, occurs outside formal institutions, often through self-directed or incidental activities, and can be intentional or unplanned~\cite{golding_informal_2009,johnson_formal_2022}. Defined as enjoyable and satisfying knowledge acquisition~\cite{bahrin_enjoyment_2021,zaibon_user_2015}, it has been shown to enhance engagement and retention in GBL and gamified contexts, with evidence linking workplace ``fun''~\cite{tews_does_2017} and tech-mediated activities~\cite{lee_comparing_2015,pendit_enjoyable_2016,leftheriotis_gamifying_2017} to greater motivation and skill development. In summary, whereas formal learning offers structure and credentialing, informal learning offers flexibility, autonomy, and often heightened intrinsic motivation, making the two modes complementary along a continuum of educational experiences.

\subsection{Impact of Immersive Technology on Learning and Motivation} \label{impact_immersive}

Motivation is a foundational element in learning, particularly in robotics education, where technical complexity creates steep learning curves. Linnenbrink and Pintrich~\cite{linnenbrink_motivation_2002} argued that motivation shapes both effort and the cognitive strategies learners employ, emphasizing the role of intrinsic and extrinsic factors in academic success. Similarly, Lynch~\cite{lynch_motivational_2006} highlighted motivational constructs such as goal orientation, self-efficacy, and interest as predictors of outcomes. These factors influence strategy selection and resource management, making motivation essential for designing effective robotics learning experiences, especially those using GBL or gamification to sustain engagement.

Recent advances in immersive technologies, particularly VR, have transformed GBL and gamified learning by enhancing engagement and immersion~\cite{dede_immersive_2009,mubarrat_physics-based_2024,mubarrat_evaluating_2024,mubarrat_evaluation_2020,srinivasan_biomechanical_2021}. VR's spatial navigation capabilities reduce cognitive load and support learners with lower spatial abilities~\cite{lee_learning_2014}, while fostering self-presence~\cite{ratan_self-presence_2013} and embodied cognition, improving retention and outcomes~\cite{shin_role_2017,steed_impact_2016}. Complementary findings show immersive extended reality (XR) exergames boosted motivation, flow, and performance~\cite{karaosmanoglu2024born}, and social VR deepened presence through embodied interactions~\cite{freeman2021body,freeman2022working}. Adding haptic feedback further enhanced learning by delivering tactile and kinesthetic cues~\cite{bonani2018my,harvey_comparison_2021,mubarrat_physics-based_2024,skola_progressive_2019}, improving motor skills~\cite{grant_audiohaptic_2019} and training outcomes~\cite{harvey_comparison_2021,skola_progressive_2019}. Beyond education, physically collaborative haptic-like guidance dramatically improved task success for blind users~\cite{bonani2018my}, highlighting the value of multi-sensory immersion in bridging virtual and real-world applications.

\subsection{Existing Reviews of Game-based and Gamified Approaches to Robotics Learning}\label{background_reviews}

Although prior research demonstrates the promise of GBL and gamified strategies in robotics education, significant gaps remain. Existing studies, such as Chen et al.~\cite{chen_gamified_2023} and Adipat et al.~\cite{adipat_engaging_2021}, largely address general educational contexts rather than robotics-specific challenges. Even targeted reviews (e.g., Fante et al.~\cite{fante_navigating_2024}; Kasenides et al.~\cite{kasenides_edifying_2024}) and meta-analyses (Ouyang and Xu~\cite{ouyang_effects_2024}; Sailer and Homner~\cite{sailer_gamification_2020}) often aggregate diverse GBL and gamified designs, obscuring critical nuances and offering limited evidence on long-term skill retention. Broader reviews (Videnovik et al.~\cite{videnovik_game-based_2023}; Miljanovic and Bradbury~\cite{miljanovic_review_2018}) further highlight overreliance on short-term studies and a lack of robotics-specific hardware integration, while La Paglia et al.~\cite{la_paglia_gaming_2023} note minimal exploration of immersive technologies, such as VR or haptics.

\subsection{Gap Synthesis and Review Scope}\label{background_gaps}

Synthesizing the current state of research, several crucial gaps emerge that necessitate further exploration. Firstly, there is limited integrated skill coverage in existing studies, particularly regarding methodologies that simultaneously teach programming, software development, and understanding of robotic systems. Secondly, immersive technologies, such as VR and haptic feedback remain understudied despite their pedagogical potential. Thirdly, informal learning environments are largely overlooked, even though they could effectively alleviate curriculum constraints. Finally, no systematic literature review has yet been conducted that analyzes both GBL and gamified approaches comprehensively, leaving a significant gap in understanding how these methodologies comparatively contribute to robotics education. 


The primary aim of this systematic literature review is to address these gaps. By explicitly focusing on technology stacks, multimodal interfaces, informal learning scenarios, and multiple robotics-related skills, this research seeks to offer comprehensive insights and practical recommendations. This review thus promises significant contributions to both theoretical understanding and practical implementations in robotics education, paving the way for more effective, engaging, and sustainable educational practices.


\section{METHODS}\label{methods}

We conducted and reported this systematic review in alignment with the PRISMA guidelines and checklist~\cite{page_prisma_2021,page_prisma_2021-1}. Adherence to the PRISMA framework ensured a rigorous, transparent, and methodologically sound process covering the identification, screening, eligibility evaluation, and inclusion of relevant studies.

\subsection{Search Strategy}\label{methods_search}

\begin{figure*}[hbt]
\centering
\includegraphics[width=1.0\linewidth]{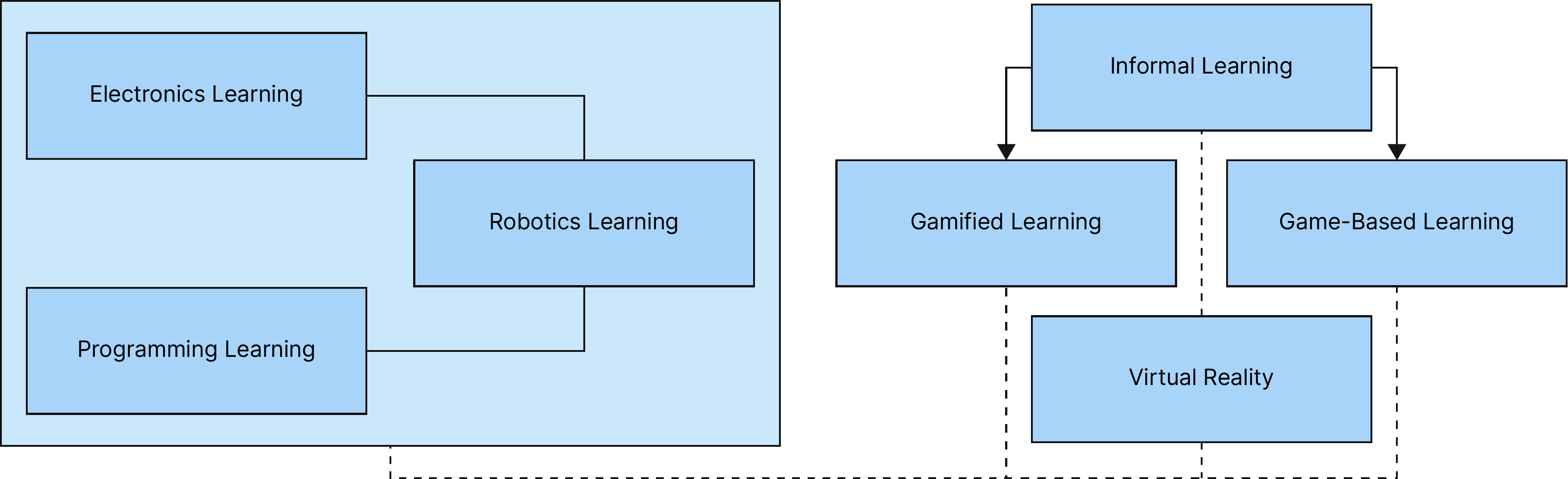}
\caption{A concept map illustrating the relationships among key concepts used for the literature review. Solid lines denote direct relationships, arrows denote hierarchy, and dashed lines denote overarching relationships.}\label{fig:concept}\Description{A conceptual diagram illustrates relationships among learning components and approaches. On the left, a blue box contains electronics learning and programming learning, both of which feed into robotics learning. On the right, informal learning connects to gamified learning and game-based learning, which both link to virtual reality.}
\end{figure*}

The literature review began with the development of a review protocol, which consisted of selecting library databases and identifying appropriate search queries. Firstly, four databases were selected based on recommendations of systematic review guides~\cite{siddaway_how_2019}: Scopus by Elsevier, ProQuest by Clarivate, IEEE Xplore by the Institute of Electrical and Electronics Engineers (IEEE), and ACM Digital Library by the Association for Computing Machinery (ACM). Using a concept map (see Figure~\ref{fig:concept}) as a guide, we then tested multiple keyword combinations and examined the results to verify their alignment with the concepts. Through this procedure, we finalized the search queries outlined in Table~\ref{tab:keywords} (detailed table available in the \href{https://osf.io/gq3a8/files/wbpqa}{Supplementary Materials}). We carried out these searches on the ``title,'' ``abstract,'' and ``keywords'' fields of the available literature in March 2025. To ensure methodological consistency and relevance to current educational practices, we only included studies published from 2014 onward. This period reflected the growing maturity of GBL and gamified learning frameworks~\cite{bahrin_enjoyment_2021,sailer_gamification_2020}, along with the integration of robotics platforms (e.g., LEGO Mindstorms EV3~\cite{rollins_beginning_2014,ucgul_history_2013,valk_lego_2014}, Arduino, and micro:bit~\cite{kalogiannakis_systematic_2021}), immersive technologies ~\cite{radianti_systematic_2020}, and computational thinking in education ~\cite{hsu_how_2018}. In addition, we limited our search to articles published in the English language with full-text availability.

\begin{table}[htbp]
\centering
\footnotesize
\caption{Keywords used for searching multiple databases (detailed table available in the \href{https://osf.io/gq3a8/files/wbpqa}{Supplementary Materials}).}
\label{tab:keywords}

\renewcommand{\arraystretch}{1.2}
\begin{tabular}{p{0.09\linewidth} p{0.85\linewidth}}
\hline
\textbf{Query} & \textbf{Keywords} \\
\hline
\textbf{S1} & robot* AND (game* OR gamifi*) AND (learn* OR teach*) \\
\textbf{S2} & ``informal learning'' AND game* AND robot* \\
\textbf{S3} & robot* AND (game* OR gamifi*) AND (learn* OR teach*) AND (``virtual reality'' OR vr) \\
\textbf{S4} & robot* AND (game* OR gamifi*) AND (learn* OR teach*) AND electronic* AND (programming OR coding) \\
\hline
\end{tabular}
\end{table}


\subsection{Eligibility criteria}\label{methods_eligibility}

To ensure the relevance and rigor of this systematic review and minimize selection bias, we explicitly defined inclusion and exclusion criteria, which can be found in the \href{https://osf.io/gq3a8/files/vbj5e}{Supplementary Materials}.

\subsection{Screening and Synthesis} \label{methods_screening}

The total number of articles retrieved from database searches is reported in the \href{https://osf.io/gq3a8/files/wbpqa}{Supplementary Materials}. After removing duplicates, we screened titles, abstracts, and keywords against the eligibility criteria. From the final set of studies on GBL and gamified approaches to teaching robotics and related skills, we prepared individual summaries and compiled them into a study matrix (see \href{https://osf.io/gq3a8/files/qzk3y}{Supplementary Materials}). We then thematically coded data through an iterative process: open coding to identify emerging concepts, followed by axial coding to cluster related ideas. Codes were inductively derived yet informed by established frameworks in GBL, gamification, and robotics education.

When studies did not explicitly classify the instructional approach (i.e., GBL or gamified), we applied definitions from Deterding et al.~\cite{deterding_game_2011}, Sailer and Homner~\cite{sailer_gamification_2020}, and Bahrin et al.~\cite{bahrin_enjoyment_2021}. Similarly, we determined learning context (i.e., formal or informal) using criteria from Johnson and Majewska~\cite{johnson_formal_2022}, Bahrin et al.~\cite{bahrin_enjoyment_2021}, Zaibon et al.~\cite{zaibon_user_2015}, and Golding et al.~\cite{golding_informal_2009}, based on the intervention rather than the evaluation design.

To categorize the level of programming/software development and understanding of robotic systems skills targeted in each study, we used an adaptation of Evripidou et al.~\cite{evripidou_educational_2020} with three levels (Basic, Intermediate, Advanced). The programming/software development skills included coding, software design, and system implementation needed to operate robotic systems. In parallel, understanding of robotic systems skills involved knowledge of electronics, mechanical assembly, control systems, and the functional integration of components, such as sensors and actuators. Two raters, each with 10 years of experience in electronics, programming, and robotics, separately classified the studies, achieving very high agreement (Cohen’s $\kappa = .918$)~\cite{cohen1960coefficient, landis1977application}. This enabled standardized comparison of skill complexity across interventions. Many studies often spanned multiple skill levels, user experience ranges, modalities, and game elements. Therefore, we applied multiple classifications where appropriate.

In analyzing reported outcomes, we focused on the effects on engagement, motivation, and learning gains described in each study. This emphasis aligns with the rationale outlined in Section~\ref{impact_immersive}, recognizing that fostering motivation is not merely advantageous but essential for designing effective robotics learning experiences. It is important to note that, in this review, we interpret learner engagement as observable behaviors (e.g., time-on-task, task completion, participation) and affective indicators (e.g., interest, enjoyment) as reported in the included studies.


We then conducted a mixed-methods synthesis to integrate qualitative insights with descriptive quantitative trends. We used frequencies and cross-tabulations to identify patterns across coded variables. Complementing this, we undertook a narrative synthesis to contextualize the results, particularly for complex educational phenomena that could not be fully captured through numerical analysis. We have included additional details about our methodological approach in the \href{https://osf.io/gq3a8/files/pnfy2}{Supplementary Materials}.


\subsection{Statistical Synthesis}\label{methods_statistics}

We complemented descriptive summaries with inferential analyses. First, we constructed a $2\times2$ contingency table to compare instructional approach and learning context (approach~$\times$~context). We also examined the distribution of pedagogies across studies. Since individual studies could be associated with multiple pedagogies, we generated separate $2\times2$ tables for each pedagogy, comparing its proportion in GBL vs. gamified approaches.

Additionally, we coded each study for the use of immersive technology (none, VR, haptics, both). Due to sparse cell counts, we collapsed these categories into a binary classification: any immersion vs.\ none, and created a $2\times2$ table. We further examined whether the likelihood of reporting outcomes differed by learning context (formal vs.\ informal) and by instructional approach (gamified vs.\ GBL). We coded each study with two non-exclusive, binary outcome indicators: (i) \emph{quantifiable learning gains} (Yes/No) and (ii) \emph{affective outcomes} (Yes/No). As these outcomes could co-occur within a study, we created separate 2~$\times$~2 tables for each factor (context $\times$ outcome type and approach $\times$ outcome type).

For all $2\times2$ comparisons, we applied Pearson's $\chi^2$ tests of independence (or Fisher's exact test when any expected cell count was $< 5$), reporting Cram\'er's $V$ as the effect size. We presented proportions with Wilson 95\% confidence intervals (CIs). In addition, we computed odds ratios (OR) and risk ratios (RR) with 95\% CIs for 2 $\times$ 2 tables.

For programming/software development and robotic systems skill tiers (none, basic, intermediate, advanced), we performed global $\chi^2$ tests ($r\times c$) and assessed ordinal trends using logistic regression (linear-by-linear association), reporting the slope ($\beta$) and $p$-values. For inferential tests, each study was assigned to its highest skill tier to maintain mutual exclusivity.

Finally, we estimated a minimal exploratory logistic regression (two-sided $\alpha = 0.05$) for the probability that a study reported quantifiable learning gains (Yes/No), with predictors approach (gamified vs.\ GBL), context (formal vs.\ informal), immersion (any immersive technology vs.\ none), and programming/software development and robotics systems skill tiers (none = 0, basic = 1, intermediate = 2, advanced = 3). We report adjusted ORs with 95\% CIs, $p$-values, log-likelihood, Akaike Information Criterion (AIC), and McFadden's pseudo-$R^2$. Analyses were conducted in Python using the \texttt{pandas}, \texttt{scipy}, and  \texttt{statsmodels} libraries.

\subsection{Risk of Bias}\label{methods_bias}


To address the heterogeneity of study designs in our corpus, we first assessed methodological quality using the Mixed Methods Appraisal Tool (MMAT)~\cite{hong2018mixed}. MMAT provides a structured appraisal and is widely used in systematic reviews for evaluating overall study quality.

For the subset of quantitative non-randomized studies, we then evaluated risk of bias using ROBINS-I~\cite{sterne_robins-i_2016}. No randomized controlled trials were present in the corpus, so RoB 2~\cite{sterne_rob_2019} was not required.


Using the MMAT, we drew on the tool's five study-type categories (qualitative research, randomized controlled trials, quantitative non-randomized studies, quantitative descriptive studies, and mixed-methods designs). MMAT's two initial screening questions were used to determine whether a study could be appraised. Two raters separately classified the studies, achieving very high agreement (category level Cohen's $\kappa = .980$). Eleven studies failed both screening questions and therefore were not further appraised, but remained included in the systematic review. For studies that passed screening, the five MMAT criteria corresponding to each study type were applied (very high agreement; criterion level Cohen’s $\kappa = .877$). MMAT ratings were used to appraise methodological quality and to contextualize evidence strength.


For quantitative non-randomized studies, we applied the ROBINS-I ~\cite{sterne_robins-i_2016} to evaluate risk of bias. Two reviewers independently assessed each study across seven domains: confounding (D1), selection of participants (D2), classification of interventions (D3), deviations from intended interventions (D4), missing data (D5), measurement of outcomes (D6), and selection of the reported result (D7). Judgments followed the standard four-level scale (``Low,'' ``Moderate,'' ``Serious,'' and ``Critical''), with the overall rating determined by the worst domain (``worst domain'' rule)~\cite{sterne_robins-i_2016}. We achieved very high agreement (overall rating Cohen's $\kappa = .826$) and resolved disagreements through discussion. We adhered to PRISMA guidelines~\cite{page_prisma_2021,page_prisma_2021-1} for the protocol and reporting.

\section{RESULTS}\label{results}

\subsection{Study Characteristics}\label{results_studyChars}

\begin{figure}[hbt]
\centering
\includegraphics[width=1.0\linewidth]{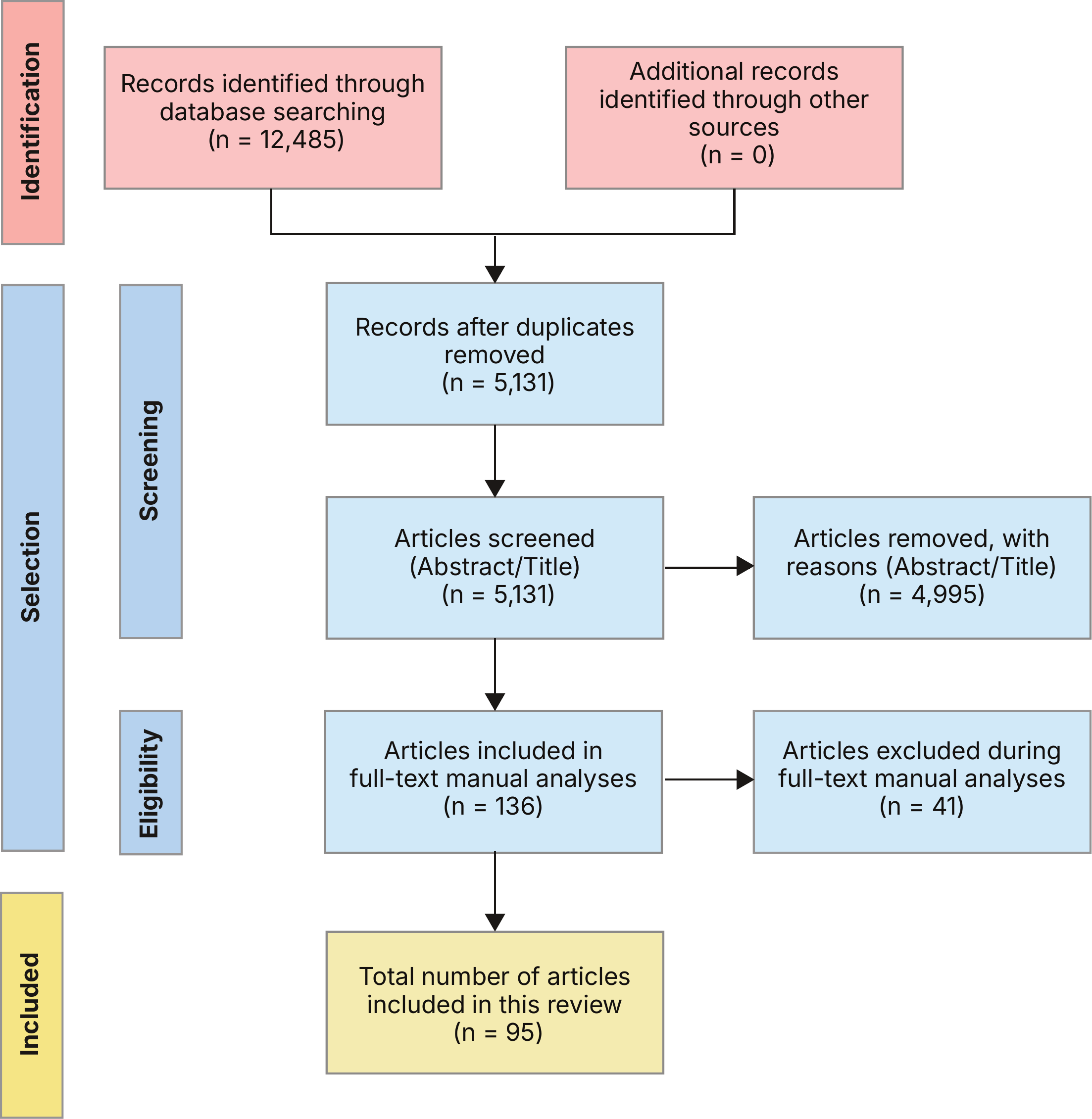}
\caption{Flowchart of the PRISMA-based selection process.}\label{fig:flowchart}\Description{A flowchart illustrates the article selection process for the review, divided into four stages. Identification: 12,485 records found through database searches, none from other sources. Screening: after removing duplicates, 5,131 records were screened by title and abstract, with 4,995 removed. Eligibility: 136 articles underwent full-text analysis, and 41 were excluded. Included: 95 articles were retained for the review.}
\end{figure}

\begin{figure}[hbt]
\centering
\includegraphics[width=1.0\linewidth]{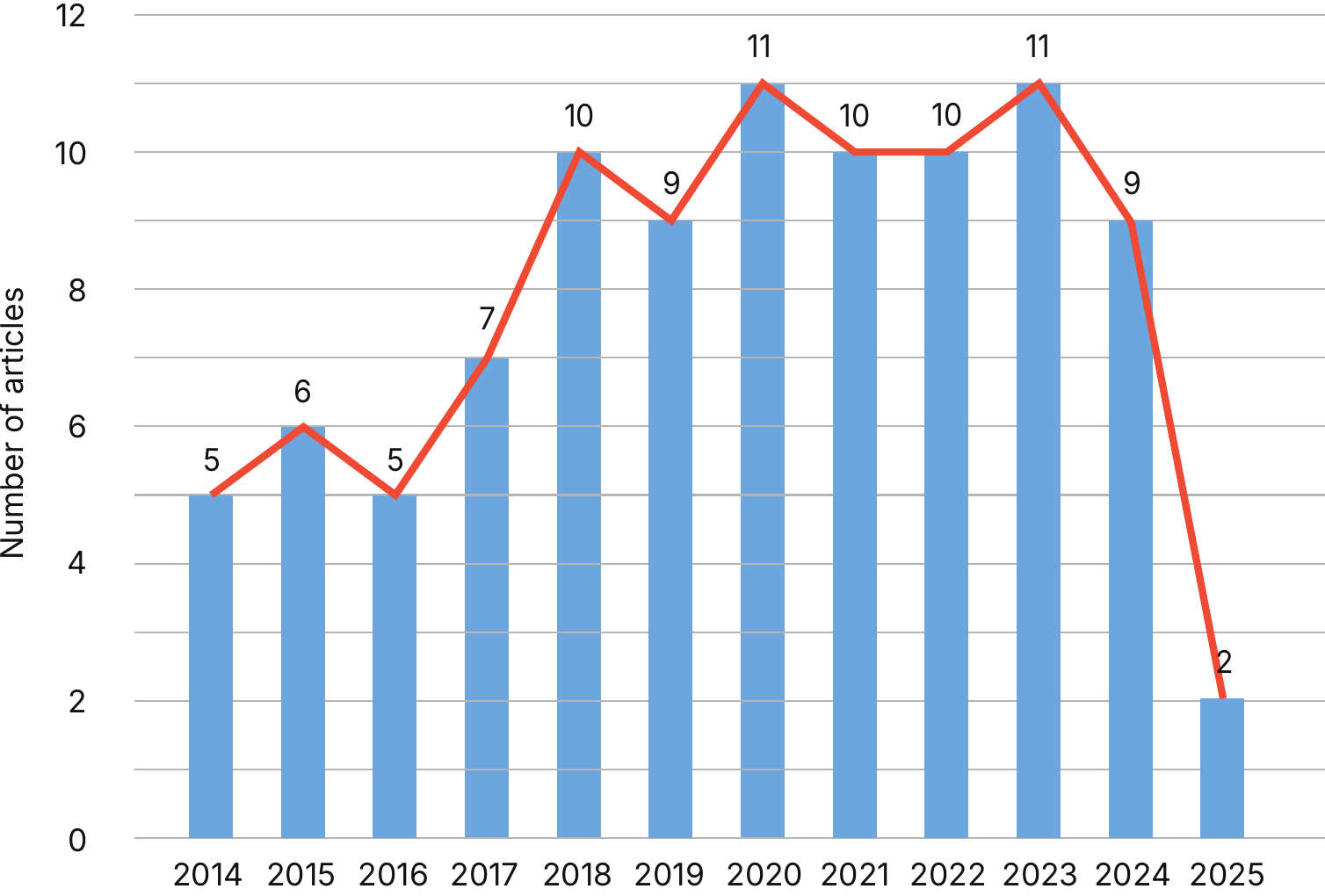}
\caption{Distribution of the publication years of the reviewed articles.}\label{fig:year}\Description{A bar graph shows the number of articles published annually from 2014 to 2025, with a red trend line connecting the bars. The total number of published articles was 5, 6, 5, 7, 10, 9, 11, 11, 10, 11, 9, and 2 in 2014, 2015, 2016, 2017, 2018, 2019, 2020, 2021, 2022, 2023, 2024, and 2025, respectively.}
\end{figure}

The information flowchart detailing the number of articles identified, screened, processed, and excluded during the manual analysis is presented in Figure~\ref{fig:flowchart}. In addition, the annual publication trends concerning GBL and gamified approaches to teaching robotics and associated skills are illustrated in Figure~\ref{fig:year}. In the following subsections, we provide an overview of the 95 studies included in the final synthesis and aim to address \textbf{RQ1}: What game-based and gamified approaches have been employed to teach robotics and associated skills (i.e., programming, software development, and understanding of robotic systems), and what trends can be observed across the core study features? The study summary matrix (available in the \href{https://osf.io/gq3a8/files/qzk3y}{Supplementary Materials}; simplified in Table \ref{tab:summary}) lists details on each study's characteristics. 


\subsubsection{Approach Types and Learning Contexts.}\label{results_studyChars_approachContext}

\begin{figure}[hbt]
\centering
\includegraphics[width=1.0\linewidth]{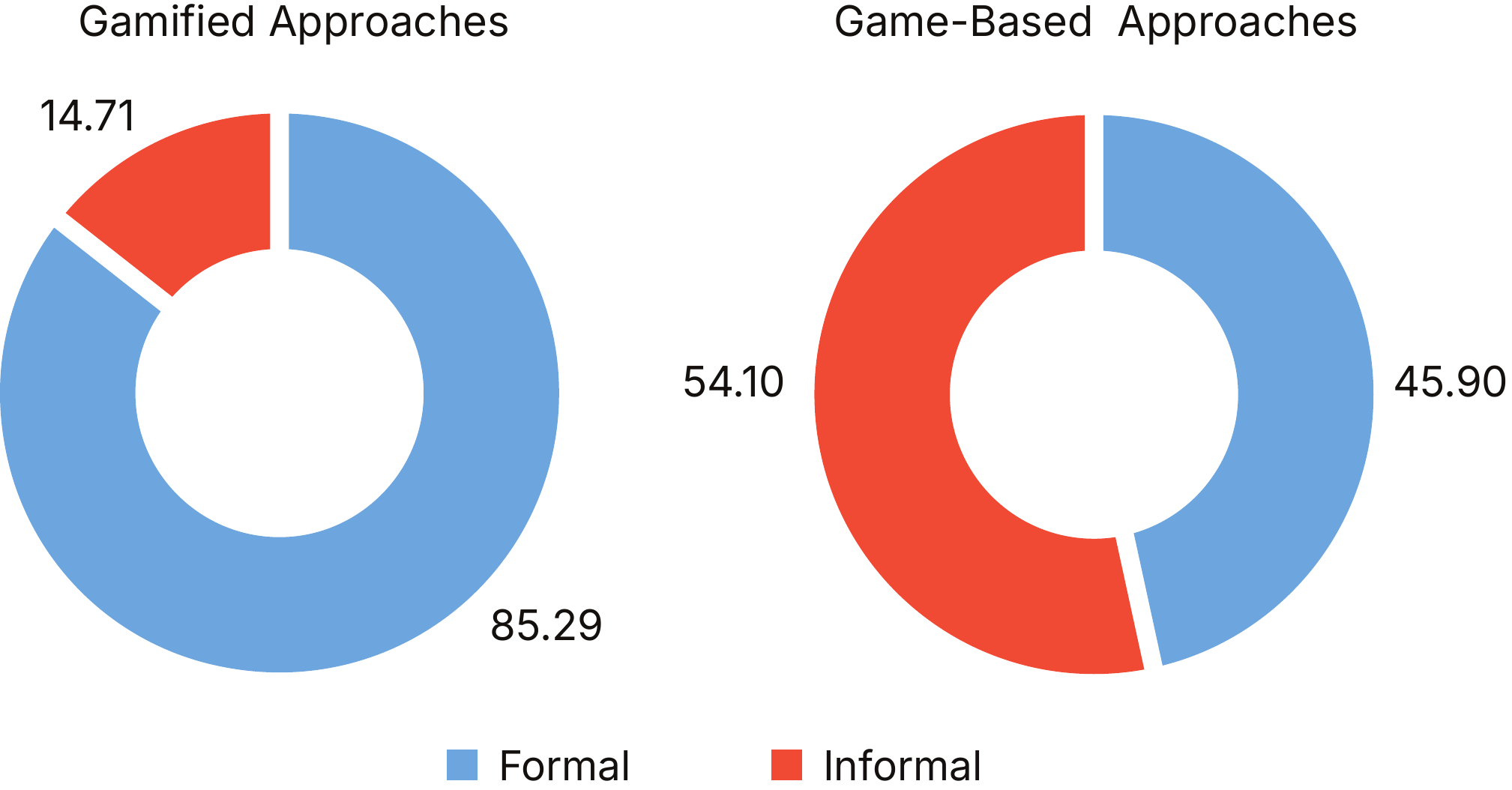}
\caption{Learning contexts utilized across different approach types in the reviewed studies (\% of studies).}\label{fig:approach_x_context}\Description{Two donut charts compare learning contexts for gamified and game-based approaches. The left chart, Gamified Approaches, shows 85.29\% formal and 14.71\% informal. The right chart, Game-Based Approaches, shows 45.90\% formal and 54.10\% informal.}
\end{figure}

Gamification was overwhelmingly tied to formal contexts (85.29\%; 95\% CI [69.87\%, 93.55\%]; see Figure~\ref{fig:approach_x_context}) and only rarely in informal environments (14.71\%; 95\% CI [6.45\%, 30.12\%]; see Figure~\ref{fig:approach_x_context}). In contrast, GBL approaches were more evenly distributed across formal (45.90\%; 95\% CI [34.01\%, 58.28\%]; see Figure~\ref{fig:approach_x_context}) and informal (54.10\%; 95\% CI [41.72\%, 65.99\%]; see Figure~\ref{fig:approach_x_context}) settings. This overall association between approach and context was statistically significant. with $\chi^2(1) = 14.16, p < 0.001$, Cramer's V $= 0.37$, OR = 6.84 (95\% CI [2.33, 20.02]), and RR = 1.86 (95\% CI [1.37, 2.52]).

\begin{figure}[hbt]
\centering
\includegraphics[width=1.0\linewidth]{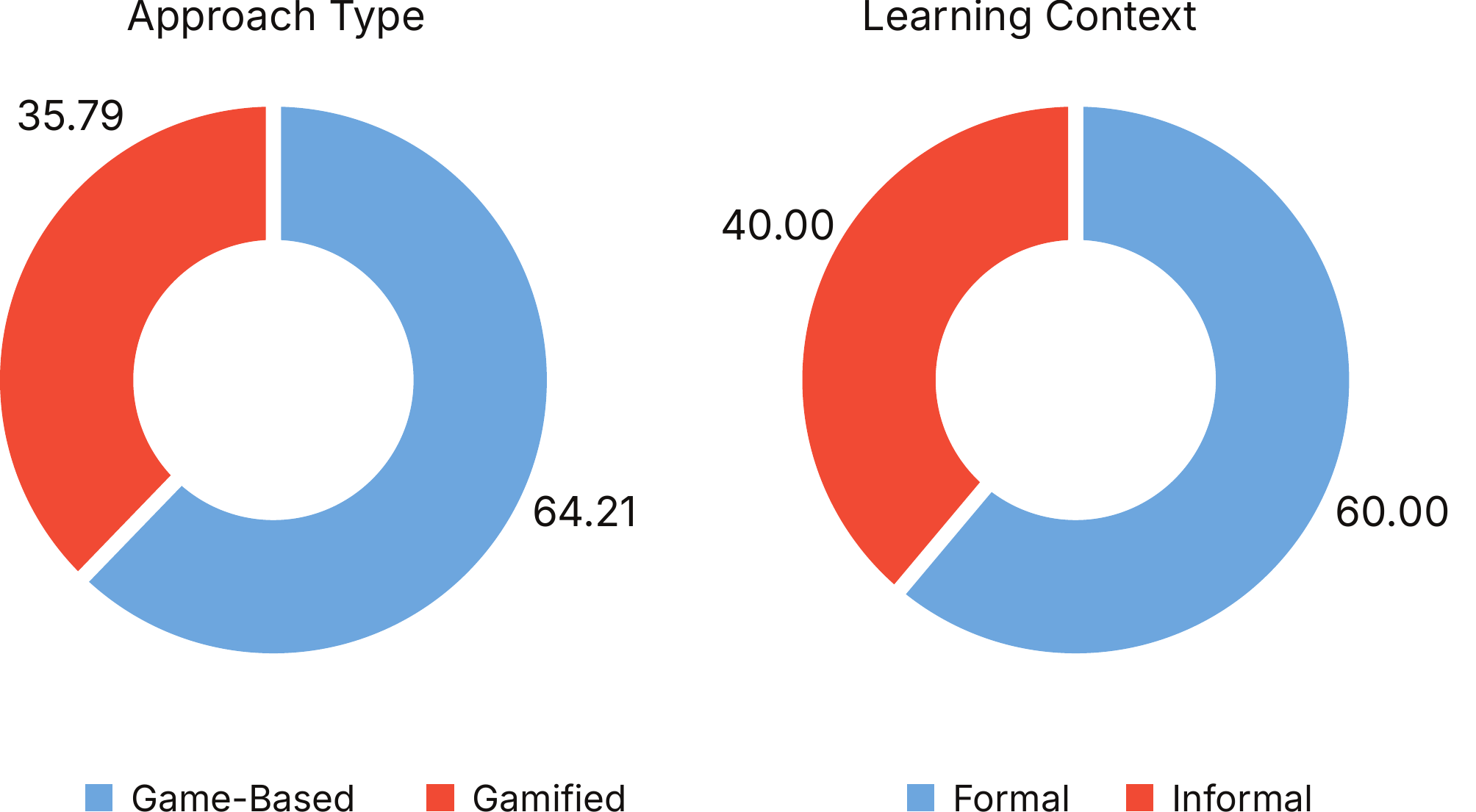}
\caption{Left: approach types utilized across the reviewed studies (\% of studies). Right: Learning contexts across the reviewed studies (\% of studies).}\label{fig:approach}\label{fig:context}\Description{Two donut charts compare distribution by approach type and learning context. The left chart, titled Approach Type, shows 64.21\% of studies as game-based and 35.79\% as gamified. The right chart, titled Learning Context, shows 60.00\% of studies in formal settings and 40.00\% in informal settings.}
\end{figure}

Analysis of the 95 included studies revealed a clear preference for fully fledged GBL environments over gamification (see Figure~\ref{fig:approach} left). Similarly, interventions were more frequently situated in formal educational contexts (see Figure~\ref{fig:context} right). For more details on various games, game elements, and learning settings used, see the \href{https://osf.io/gq3a8/files/esj8h}{Supplementary Materials}.

\subsubsection{Pedagogical Models.}\label{results_studyChars_pedagogy}

Our review of 95 studies identified three dominant pedagogical models in GBL and gamified robotics learning: experiential learning, project-based learning (PBL), and constructivist learning. Kolb’s experiential learning theory \cite{kolb_experiential_2015} describes learning as a cyclical process of experiencing, reflecting, conceptualizing, and experimenting to transform experience into knowledge. PBL is a student-centered approach that engages learners in solving real-world problems through collaborative, interdisciplinary projects that integrate knowledge and skills \cite{jalinus_seven_2017,lin_applying_2022,larmer_setting_2015,markham_project_2012}. Constructivist learning theory (originally developed by Piaget~\cite{piaget_development_1972,piaget_understand_1973}) in robotics education holds that learners build knowledge by making and manipulating programmable artifacts, promoting exploratory, iterative, and personally meaningful problem solving \cite{papert_mindstorms_2020,alimisis_educational_2013,druin_robots_2000}.

Constructivist pedagogy was more common in GBL than in gamified studies (GBL: 49.18\%, 95\% CI [37.06\%, 61.40\%]; gamified: 25.71\%, 95\% CI [14.60\%, 43.12\%]). This association was statistically significant, with $\chi^2(1) = 4.65, p = .031$, Cramer's V $= 0.20$, OR = 0.37 (95\% CI [0.15, 0.93]), and RR = 0.54 (95\% CI [0.29, 0.99]).

Experiential pedagogy did not differ by approach (GBL: 60.66\%, 95\% CI [48.12\%, 71.93\%]; gamified: 54.29\%, 95\% CI [39.45\%, 71.11\%]). The association was not significant, with $\chi^2(1) = 0.21, p = .650$, Cramer's V $= 0.00$, OR = 0.82 (95\% CI [0.35, 1.92]), and RR = 0.92 (95\% CI [0.64, 1.32]).

PBL was more frequently used in gamified than in GBL studies (GBL: 45.90\%, 95\% CI [34.01\%, 58.28\%]; gamified: 71.43\%, 95\% CI [56.88\%, 85.40\%]). This association was statistically significant, with $\chi^2(1) = 6.76, p = .009$, Cramer's V $= 0.25$, OR = 3.27 (95\% CI [1.31, 8.16]), and RR = 1.60 (95\% CI [1.14, 2.25]).

\begin{figure*}[hbt]
\centering
\includegraphics[width=1.0\linewidth]{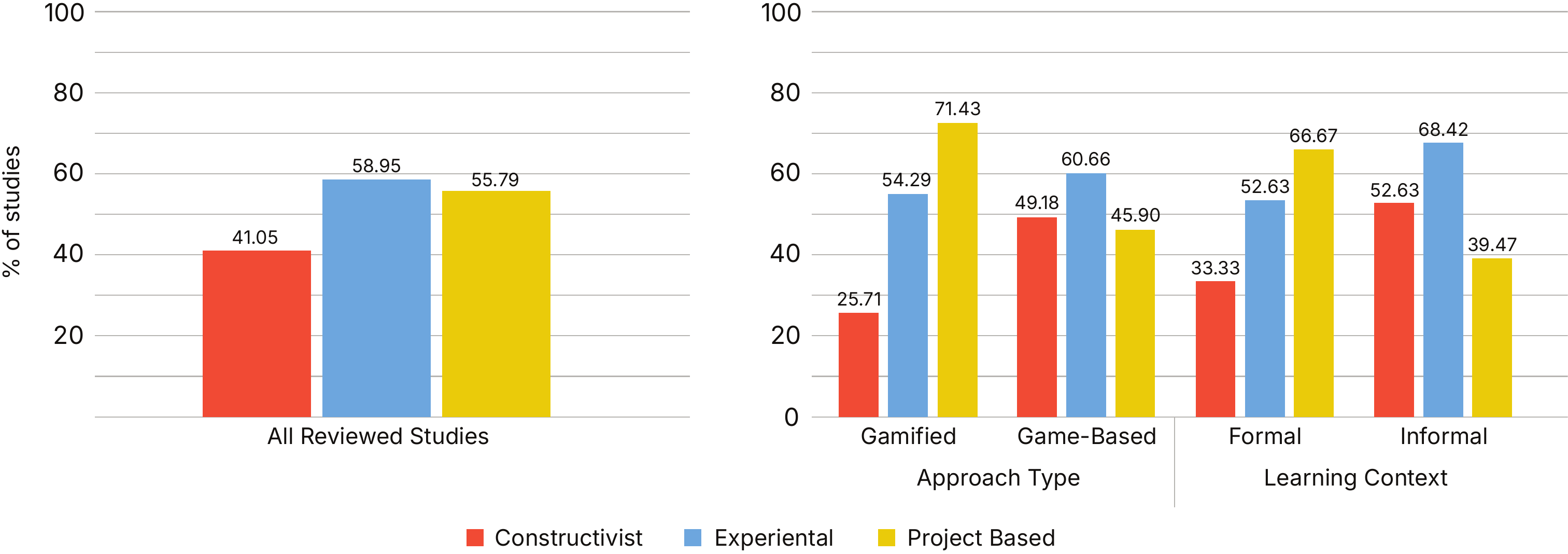}
\caption{Left: pedagogical models used across the reviewed studies (\% of studies). Right: Comparison of different pedagogical models across approach types and learning contexts (\% of studies).}\label{fig:pedagogy}\label{fig:pedagogy_x_approach_x_context}\Description{Two bar charts compare pedagogical approaches across studies. The left chart shows overall percentages: Constructivist 41.05\%, Experiential 58.95\%, and Project-Based 55.79\%. The right chart breaks these down by approach type and learning context. For gamified approaches: Constructivist 25.71\%, Experiential 54.29\%, Project-Based 71.43\%. For game-based: Constructivist 49.18\%, Experiential 60.66\%, Project-Based 45.90\%. In formal contexts: Constructivist 33.33\%, Experiential 52.63\%, Project-Based 66.67\%. In informal contexts: Constructivist 52.63\%, Experiential 68.42\%, Project-Based 39.47\%.”}
\end{figure*}

Across the corpus, the analysis revealed a strong preference for experiential designs, often combined with PBL and, to a lesser extent, constructivist pedagogy (see Figure~\ref{fig:pedagogy} left), reflecting the prevalence of blended pedagogical models. Gamified interventions tended to prioritize PBL and experiential strategies (see Figure~\ref{fig:pedagogy_x_approach_x_context} right), while GBL studies exhibited a more balanced mix of experiential and constructivist elements, with PBL playing a secondary role (see Figure~\ref{fig:pedagogy_x_approach_x_context} right). Mapping pedagogy to context further indicated that formal classrooms lean toward PBL-driven designs, whereas informal settings favor experiential approaches and included higher proportions of constructivist elements (see Figure~\ref{fig:pedagogy_x_approach_x_context} right).

\subsubsection{Levels of Skills Taught.}\label{results_studyChars_skills}


In terms of programming/software development skill levels, GBL studies reported 63.93\% basic (95\% CI [51.39\%, 74.83\%]), 19.67\% intermediate (95\% CI [11.63\%, 31.31\%]), and 14.75\% advanced (95\% CI [7.96\%, 25.72\%]), whereas gamified studies reported 35.29\% basic (95\% CI [21.49\%, 52.09\%]), 50.00\% intermediate (95\% CI [34.07\%, 65.93\%]), and 14.71\% advanced (95\% CI [6.45\%, 30.13\%]). The distribution of programming/software development skill levels differed significantly between GBL and gamified studies ($\chi^2(2) = 10.47$, $p = .015$, Cramer’s $V = 0.28$), with gamified approaches skewing toward higher tiers (logistic trend $\beta = 0.57$, $p = .049$).

In contrast, robotics systems skill levels did not differ by approach. GBL studies reported 32.79\% basic (95\% CI [22.34\%, 45.28\%]), 19.67\% intermediate (95\% CI [11.63\%, 31.31\%]), and 4.92\% advanced (95\% CI [1.69\%, 13.50\%]), while gamified studies reported 29.41\% basic (95\% CI [16.83\%, 46.17\%]), 29.41\% intermediate (95\% CI [16.83\%, 46.17\%]), and 5.88\% advanced (95\% CI [1.63\%, 19.10\%]). No significant differences were observed ($\chi^2(2) = 1.30$, $p = .728$, Cramer’s $V = 0.00$), and the logistic trend test indicated no trend (logistic trend $\beta = 0.22$, $p = .334$).

\begin{figure*}[hbt]
\centering
\includegraphics[width=1.0\linewidth]{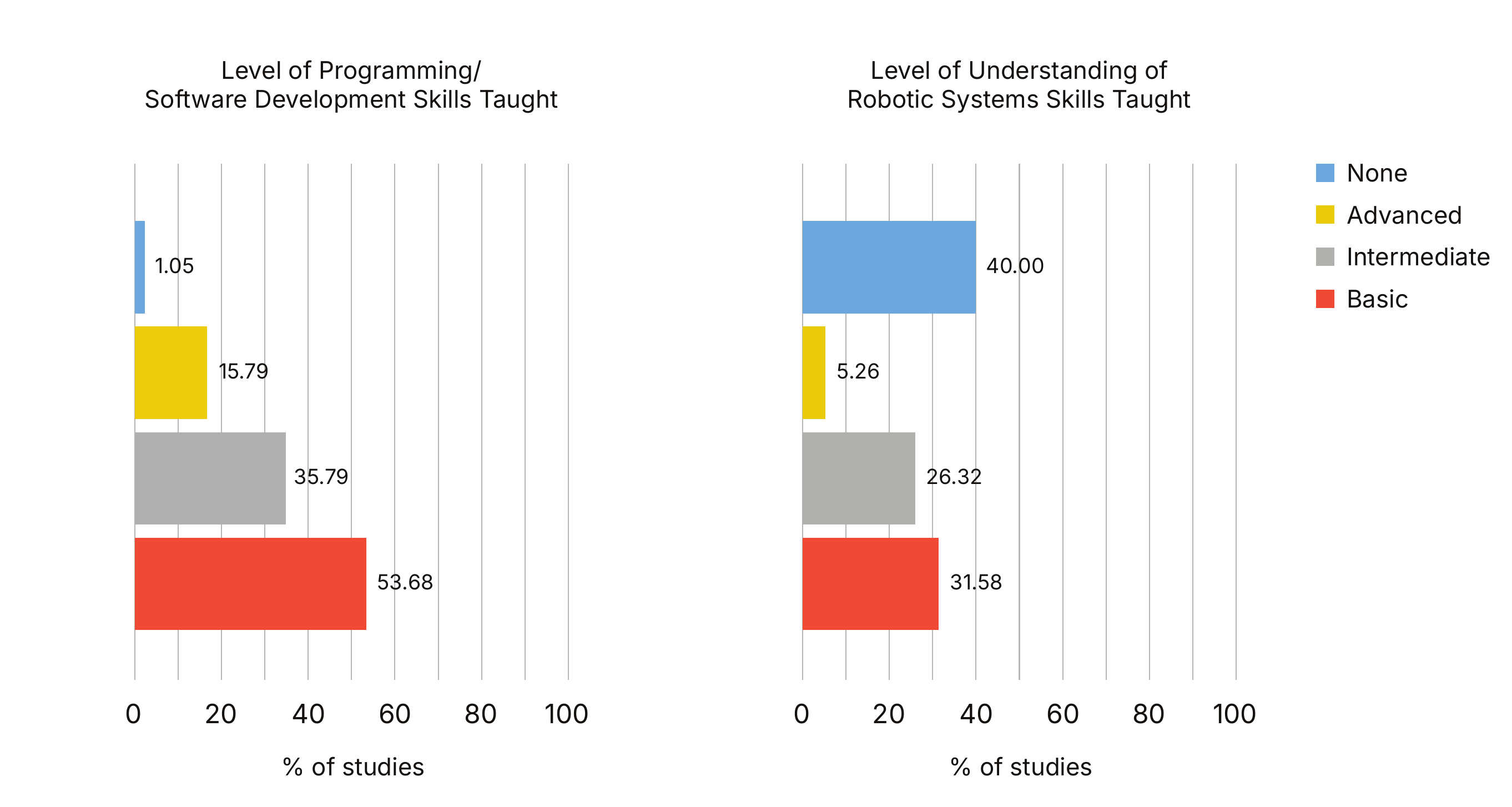}
\caption{Levels of programming/software development and understanding of robotic systems skills taught across all reviewed studies (\% of studies).}\label{fig:prog_elec_01}\Description{Two bar charts show skill levels taught in reviewed studies. The left chart, Programming/Software Development Skills, reports 1.05\% none, 15.79\% advanced, 35.79\% intermediate, and 53.68\% basic. The right chart, Understanding of Robotic Systems, shows 40.00\% none, 5.26\% advanced, 26.32\% intermediate, and 31.58\% basic.”}
\end{figure*}

\begin{figure*}[hbt]
\centering
\includegraphics[width=1.0\linewidth]{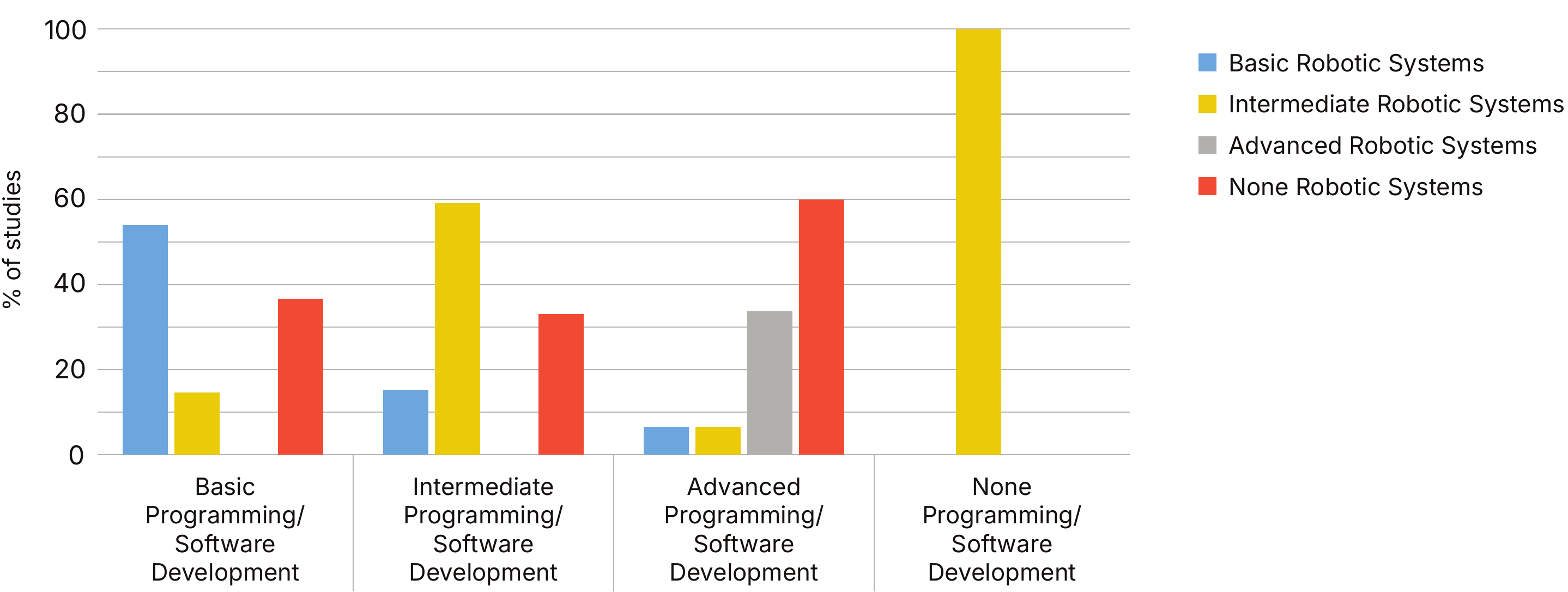}
\caption{Levels of programming/software development skills vs. understanding of robotic systems skills taught across all reviewed studies (\% of studies).}\label{fig:prog_elec_02}\Description{A bar chart compares programming skill levels with robotic system complexity across studies. Categories on the x-axis are basic, intermediate, advanced, and none for programming/software development. Bars show percentages for robotic system levels: basic (blue), intermediate (yellow), advanced (gray), and none (red). For basic programming: 54\% basic robotics, 14\% intermediate, 0\% advanced, 37\% none. For intermediate programming: 15\% basic, 63\% intermediate, 6\% advanced, 34\% none. For advanced programming: 6\% basic, 31\% intermediate, 63\% advanced, 0\% none. For no programming: 0\% basic, 100\% intermediate, 0\% advanced, 0\% none.}
\end{figure*}

Across all studies, instruction predominantly concentrated on basic and intermediate programming/software development, with advanced coverage remaining rare (see Figure~\ref{fig:prog_elec_01}). Robotic systems skills were addressed far less frequently and mostly at lower tiers (see Figure~\ref{fig:prog_elec_01}), indicating a gap in comprehensive hardware-software integration. Overall, 58.95\% of studies targeted both programming / software development and robotic systems skills, with common alignment between programming and hardware levels (as exemplified by Kobayashi et al.~\cite{kobayashi_design_2018} and Martín~\cite{martin_open_2018}; see Figure~\ref{fig:prog_elec_02}), suggesting a tendency toward tier-matched design. However, advanced programming was frequently taught in isolation, lacking corresponding systems instruction (see Figure~\ref{fig:prog_elec_02}), which points to a disconnect in preparing learners for holistic, high-level robotics competencies. See the \href{https://osf.io/gq3a8/files/7c3sj}{Supplementary Materials} for a detailed table on the types of programming / software development and understanding of robotic systems skills by level taught across included studies.

\subsubsection{Usage of Immersive Technology.}\label{results_studyChars_immersive}

\begin{figure}[hbt]
\centering
\includegraphics[width=0.92\linewidth]{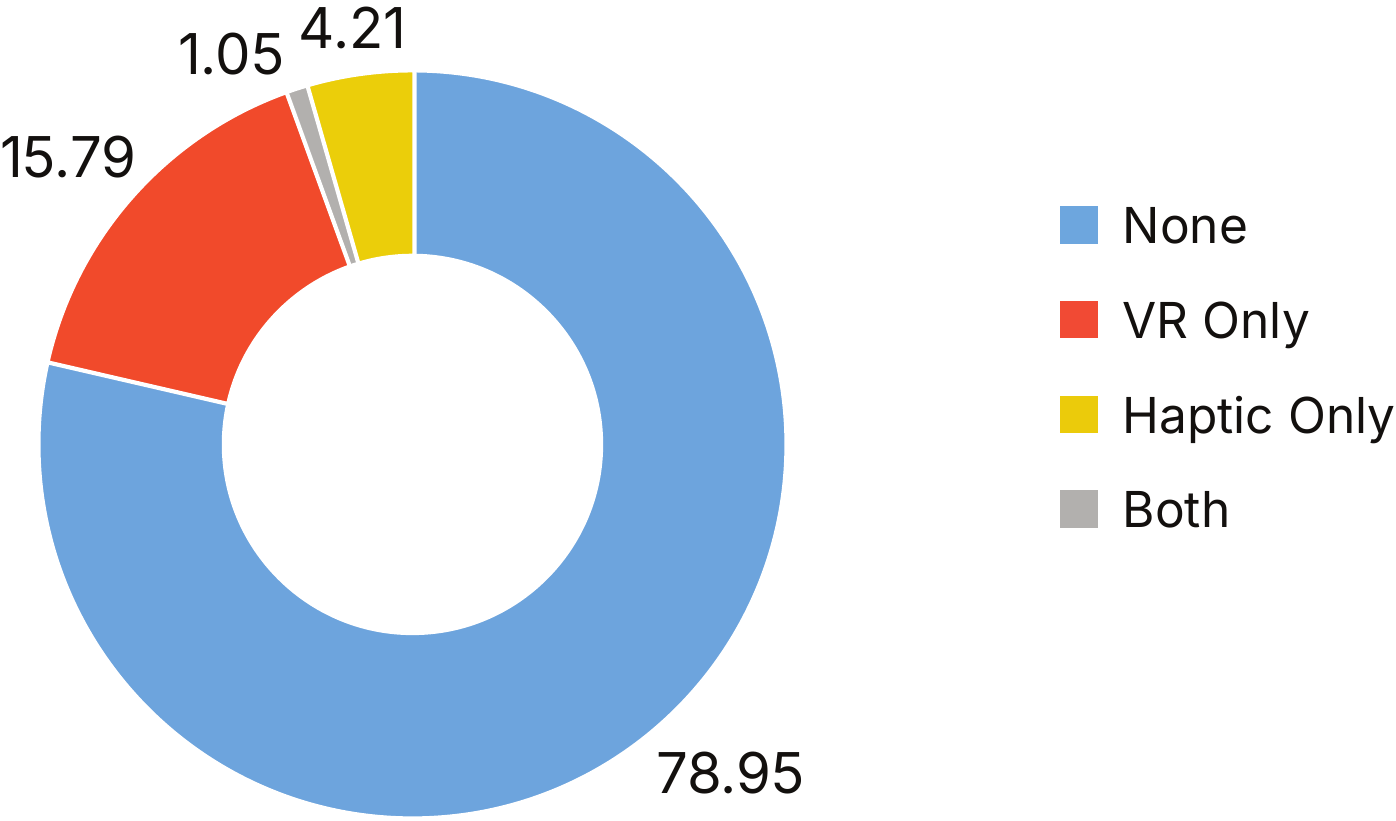}
\caption{Virtual reality (VR) and haptic technology usage across all reviewed studies (\% of studies).}\label{fig:VR_haptic}\Description{A donut chart shows the distribution of immersive technology use in studies. Most studies (78.95\%) used none. Virtual Reality (VR) only accounts for 15.79\%, haptic only for 4.21\%, and both VR and haptic for 1.05\%.}
\end{figure}

Most studies relied on conventional screen-based or physical-robot interactions (see Figure~\ref{fig:VR_haptic}), with immersive technologies playing only a minor role. VR emerged as the predominant alternative to traditional setups (see Figure~\ref{fig:VR_haptic}), likely driven by the increasing accessibility of head-mounted displays and development toolkits. Haptic feedback, whether standalone or combined with VR~\cite{peterson_teaching_2021,peterson_evaluating_2022,vassigh_performance-driven_2023} (see Figure~\ref{fig:VR_haptic}), was rarely incorporated, signaling a significant underutilization of tactile interaction despite its potential to enhance embodied learning experiences.

Immersion was similar across approaches ($\chi^2(1) = 0.20$, $p = 0.658$; Cramér's $V \approx 0.00$). In gamified studies, 23.53\% used immersive technology (VR/haptics/both; 95\% CI [12.44\%, 40.00\%]) versus 19.67\% in GBL (95\% CI [11.63\%, 31.31\%]). The OR for any immersion (gamified vs.\ GBL) was $\approx 1.26$ (95\% CI [0.46, 3.46]), indicating a small, non-significant tendency for gamified studies to use immersive technologies.

\subsubsection{Experience Level of Target Users.}\label{results_studyChars_experience}

\begin{figure}[hbt]
\centering
\includegraphics[width=1.0\linewidth]{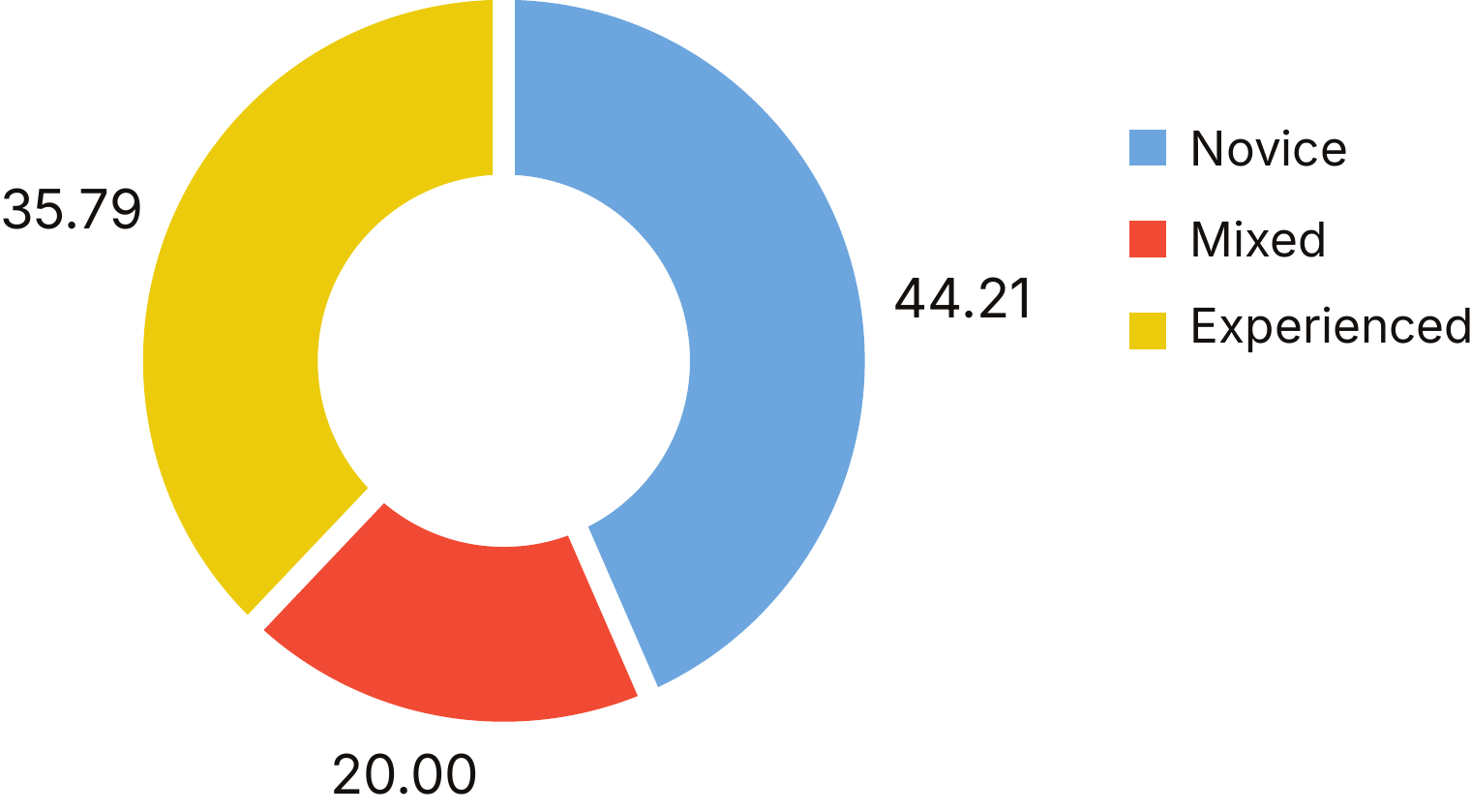}
\caption{Experience level of target users across the reviewed studies (\% of studies).}\label{fig:exp_level}\Description{A donut chart shows participant experience levels in studies. Novice participants account for 44.21\%, experienced participants for 35.79\%, and mixed experience groups for 20.00\%.}
\end{figure}

Learner experience levels in the reviewed studies reveal a strong orientation toward beginners (see Figure~\ref{fig:exp_level}), with many interventions tailored for individuals lacking prior exposure to robotics, programming, or electronics—such as K-12 students~\cite{bas_enhancing_2023,dahal_introductory_2023,hrbacek_step_2018,leonard_using_2016,liu_stem_2014} and first-year undergraduates~\cite{agalbato_robo_2018,gabriele_educational_2017,hamann_gamification_2018,han_skynetz_2016,kobayashi_design_2018,thanyaphongphat_game-based_2020}. A substantial subset focused on advanced cohorts (see Figure~\ref{fig:exp_level}), including senior engineering students~\cite{budiman_gamification_2025,crespo_virtual_2015,crespo_virtual_2015-1,eliza_game-d_2025,lee_learning_2019,praveena_effective_2024}, graduate students~\cite{lucan_simulator-based_2023,maurelli_blackpearl_2023,vazquez-hurtado_adapting_2022}, educators~\cite{kilic_exploring_2022,vazquez-hurtado_virtual_2024}, and industry professionals~\cite{hempe_erobotics_2015,peterson_evaluating_2022,vassigh_performance-driven_2023}. A smaller but noteworthy group of studies adopted inclusive strategies for mixed-proficiency cohorts~\cite{gharib_novel_2021,jansen_pump_2016,karageorgiou_escape_2019,kashinath_narvekar_learn_2020,leung_project_2022,orozco_gomez_ubuingenio_2023} (see Figure~\ref{fig:exp_level}), often leveraging adaptive difficulty or evaluating distinct learner groups within the same intervention, signaling a growing interest in scalable, differentiated learning designs.

\subsubsection{Intervention Types.}\label{results_studyChars_intervention}

\begin{figure*}[hbt]
\centering
\includegraphics[width=0.9\linewidth]{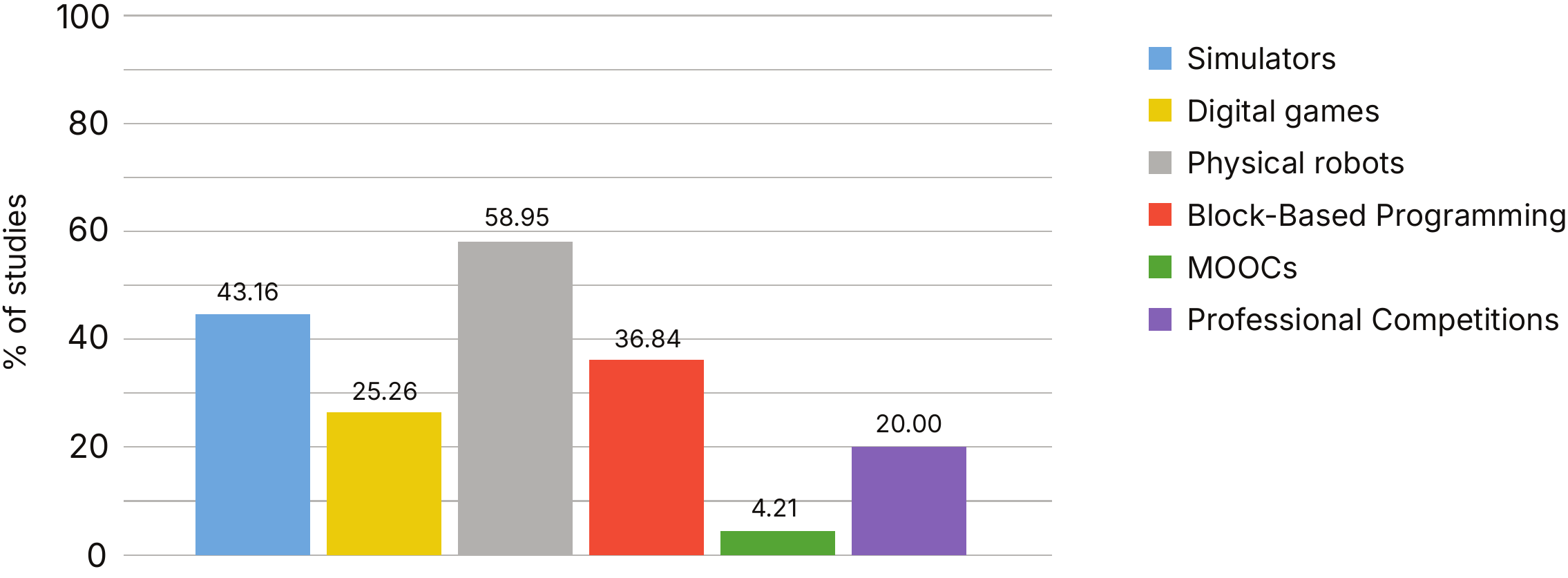}
\caption{Intervention types across all reviewed studies (\% of studies).}\label{fig:interv}\Description{A bar chart shows the percentage of studies using various intervention modalities. Physical robots appear most frequently at 58.95\%, followed by simulators at 43.16\%, block-based programming at 36.84\%, and digital games at 25.26\%. Professional competitions account for 20.00\%, and MOOCs for 4.21\%.}
\end{figure*}

Across the 95 interventions reviewed, intervention modality primarily relied on physical robots (58.95\%; see Figure~\ref{fig:interv}), typically via educational kits (e.g., LEGO Mindstorms~\cite{atmatzidou_didactical_2017,de_araujo_prorobot_2023,duraes_gaming_2015,kilic_exploring_2022,leonard_using_2016,merkouris_programming_2021,panskyi_holistic_2021,park_engaging_2013,sovic_how_2014,strnad_programming_2017,thanyaphongphat_game-based_2020,wang_codingprogramming_2019,yadagiri_blocks-based_2015}, micro:bit~\cite{stoffova_how_2023}, Thymio~\cite{giang_exploring_2020,hamann_gamification_2018}, mBots~\cite{huang_design_2017}) to support hands-on, embodied problem-solving. Nearly as prevalent were virtual simulators (43.16\%; see Figure~\ref{fig:interv}), often leveraging ROS~\cite{ates_work_2022,canas_open-source_2020,canas_ros-based_2020,costa_robotics_2016,fernandez-ruiz_automatic_2022,hamann_gamification_2018,martin_open_2018,maurelli_blackpearl_2023,roldan-alvarez_unibotics_2024,teogalbo_mixed_2023} with Gazebo~\cite{canas_open-source_2020,canas_ros-based_2020,fernandez-ruiz_automatic_2022, hamann_gamification_2018, martin_open_2018, roldan-alvarez_unibotics_2024} or Unity~\cite{asut_developing_2024,crespo_virtual_2015,crespo_virtual_2015-1,mubarrat_geobotsvr_2024, nascimento_sbotics-gamified_2021,praveena_effective_2024,vassigh_performance-driven_2023,vazquez-hurtado_virtual_2024} for immersive, scalable experiences. Self-contained digital games (e.g., \textit{Expedition Atlantis}~\cite{alfieri_case_2015}, \textit{Kodu Game Lab}~\cite{sovic_how_2014}) appeared in 25.26\% of studies (see Figure~\ref{fig:interv}), usually as conceptual scaffolds before learners progressed to simulators or hardware. Block-based visual programming featured in 36.84\% of studies (predominantly using Scratch~\cite{de_araujo_prorobot_2023,hamann_gamification_2018,kunovic_comparison_2021,mendonca_digital_2020,orozco_gomez_ubuingenio_2023,panskyi_holistic_2021,plaza_scratch_2019,plaza_stem_2019,roscoe_teaching_2014,sharma_coding_2019,wang_codingprogramming_2019} or Blockly~\cite{diaz-lauzurica_computational_2019,heljakka_gamified_2019,jaggle_effectiveness_2023,merkouris_programming_2021,moreno-vera_comparison_2018,nordin_mobile_2020,yadagiri_blocks-based_2015,zainal_primary_2018}; see Figure~\ref{fig:interv}), frequently layered over simulators and robots. Two formats were underrepresented: massive open online courses (MOOCs; 4.21\%; see Figure~\ref{fig:interv}), all delivered through the open-access RoboticsAcademy framework~\cite{canas_open-source_2020,canas_ros-based_2020,fernandez-ruiz_automatic_2022,roldan-alvarez_unibotics_2024}, and professional/semi-formal competitions (20.00\%; see Figure~\ref{fig:interv}) spanning events such as eYRC~\cite{jadhav_pbl_2021,kashinath_narvekar_learn_2020,panwar_analyzing_2020,sarkar_teaching_2020,sarkar_teaching_2022}, Zero Robotics~\cite{liu_stem_2014}, RoboFest~\cite{chung_design_2021}, and RoboMaster~\cite{leung_project_2022}. Notably, many studies adopted multimodal designs, blending two or more channels to stage learning progression. For more details on the different modalities of interventions used, refer to the \href{https://osf.io/gq3a8/files/bfe29}{Supplementary Materials}.

\subsubsection{Genre of Game or Gamification Elements.}\label{results_studyChars_genre}

\begin{figure*}[hbt]
\centering
\includegraphics[width=1.0\linewidth]{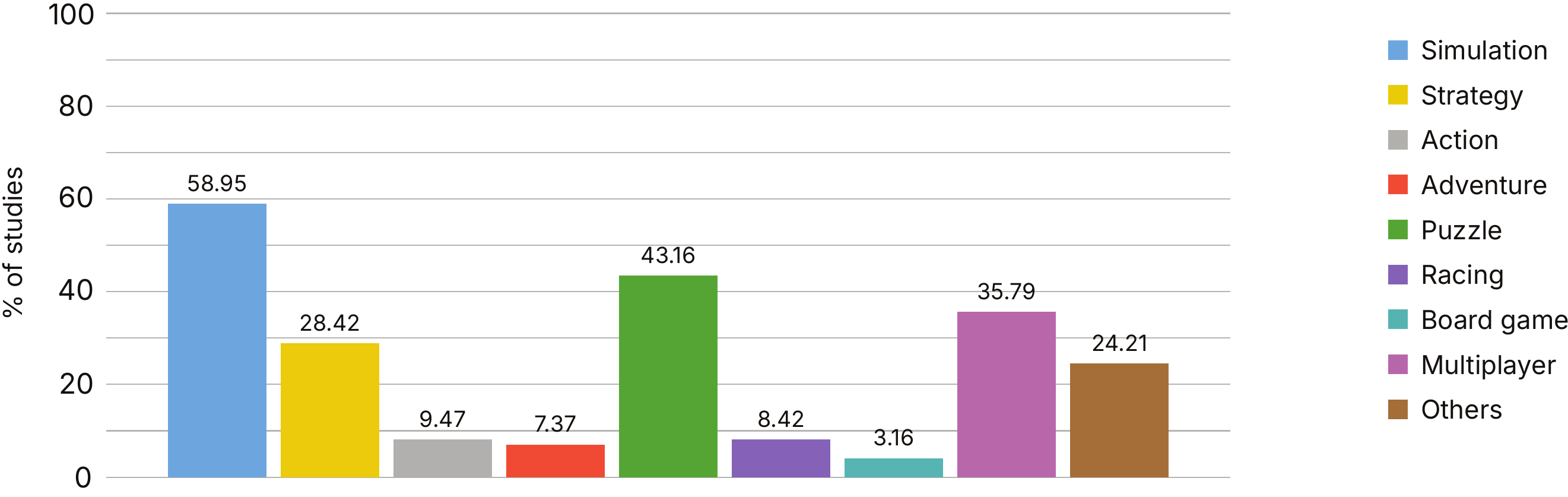}
\caption{Genre of game or gamification elements used across all reviewed studies (\% of studies).}\label{fig:genre}\Description{A bar chart shows the percentage of studies by game or gamification type. Simulation games are most common at 58.95\%, followed by puzzle games at 43.16\% and multiplayer games at 35.79\%. Strategy games account for 28.42\%, others 24.21\%, action 9.47\%, racing 8.42\%, adventure 7.37\%, and board games 3.16\%.}
\end{figure*}

\begin{figure*}[hbt]
\centering
\includegraphics[width=1.0\linewidth]{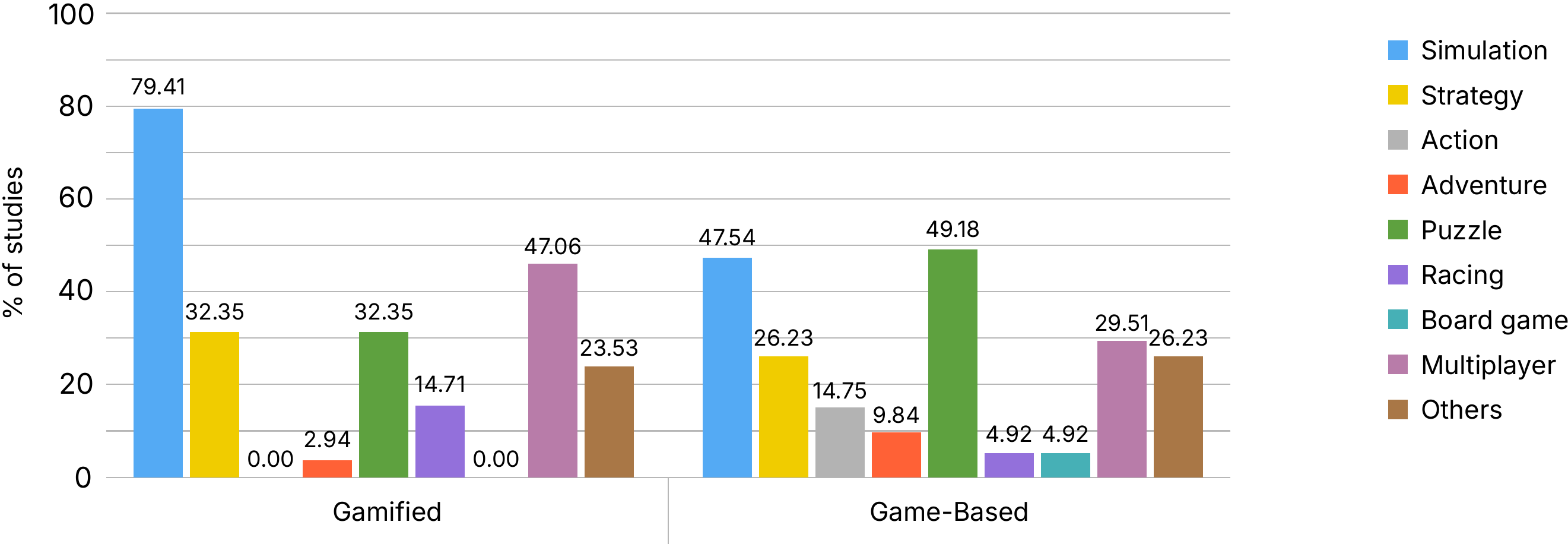}
\caption{Genre of game or gamification elements used in different approach types across all reviewed studies (\% of studies).}\label{fig:genre_x_approach}\Description{A bar chart compares game or gamification types between gamified and game-based approaches. For gamified studies: simulation 79.41\%, strategy 32.35\%, puzzle 32.35\%, multiplayer 47.06\%, others 23.53\%, racing 14.71\%, adventure 2.94\%, and none for action or board games. For game-based studies: puzzle 49.18\%, simulation 47.54\%, multiplayer 29.51\%, others 26.23\%, strategy 26.23\%, action 14.75\%, adventure 9.84\%, and both racing and board games at 4.92\%.}
\end{figure*}

Across the reviewed studies, game genres that promoted problem-solving and collaboration dominated robotics education (see Figure~\ref{fig:genre}), with simulation games emerging as the most prevalent, followed by puzzles and multiplayer formats. Gamified interventions tended to prioritize simulation and incorporate multiplayer elements (see Figure~\ref{fig:genre_x_approach}), reflecting a focus on realism and social engagement. In contrast, GBL leaned more toward puzzle-based designs (see Figure~\ref{fig:genre_x_approach}), suggesting an emphasis on cognitive challenge and structured problem-solving. Board-game metaphors were largely absent, appearing only sporadically in GBL for simple mechanics such as point tracking or turn-taking (see Figure~\ref{fig:genre_x_approach}), indicating limited exploration of analog game principles in digital or hybrid contexts.

\subsection{Effects on Learning Outcomes and Motivation}\label{results_outcomes}

The systematic literature review reveals significant insights into the effects of GBL and gamified approaches on learning outcomes in robotics education. This section synthesizes the positive outcomes of these approaches and aims to address \textbf{RQ2}: How do the different types of core study features differ in their effects on learning outcomes, motivation, and other affective outcomes?

In the exploratory logistic regression predicting whether a study reported quantifiable learning gains (Model fit: McFadden's pseudo-$R^2 = 0.091$, AIC $= 118.32$), learning context (formal vs.\ informal) was a significant positive predictor (OR = 3.17, 95\% CI [1.00, 10.08], $p = .049$). Immersion (any vs.\ none) showed a suggestive positive association (OR = 3.00, 95\% CI [0.83, 10.86], $p = .094$). Programming/software development skill tier had a negative, non-significant association (OR = 0.45, 95\% CI [0.20, 1.00], $p = .051$). Raw counts and full output tables are provided in the \href{https://osf.io/gq3a8/files/4y8hn}{Supplementary Materials}.

\subsubsection{Comparative Effectiveness of Game-based and Gamified Approaches.}\label{results_outcomes_approach}

Across the corpus, quantifiable learning gains were comparable between GBL and gamified approaches (GBL: 29.41\%, 95\% CI [18.62\%, 48.05\%]; gamified: 31.15\%, 95\% CI [20.97\%, 43.61\%]; $\chi^2(1) = 0.03$, $p = .860$, Cramér’s $V \approx 0.00$, OR = 0.92 (95\% CI [0.37, 2.30]), RR = 0.94 (95\% CI [0.50, 1.79]); see Table~\ref{tab:outcome_approach} for examples). Similarly, affective outcomes did not differ significantly by approach (GBL: 52.94\%, 95\% CI [38.21\%, 69.55\%]; gamified: 57.38\%, 95\% CI [44.93\%, 69.04\%]; $\chi^2(1) = 0.17$, $p = .676$; Cramér’s $V \approx 0.00$, OR = 0.84 (95\% CI [0.36, 1.94]), RR = 0.92 (95\% CI [0.63, 1.35]); see Table~\ref{tab:outcome_approach} for examples and the \href{https://osf.io/gq3a8/files/4y8hn}{Supplementary Materials} for details).

\begin{table}[htbp]
\centering
\footnotesize
\caption{The summary of learning outcomes of game-based and gamified approaches across the reviewed studies.}
\label{tab:outcome_approach}

\renewcommand{\arraystretch}{1.2}
\begin{tabular}{p{0.15\linewidth} p{0.22\linewidth} p{0.12\linewidth} p{0.35\linewidth}}
\hline
\textbf{Approach type} &
\textbf{Positive outcome cluster} &
\textbf{References} &
\textbf{Detailed outcomes} \\
\hline
Game-based &
Heightened affective engagement and early-stage conceptual understanding &
\cite{karageorgiou_escape_2019} &
Edison-robot programming was ``very easy'' and ``really enjoyable'' \\
&
&
\cite{peterson_evaluating_2022} &
Increased (\textgreater{} 1.90\%) self-reported confidence, motivation, interest, and task-completion efficiency in robotics education compared to traditional educational methods \\
&
Other quantitative learning gains &
\cite{thanyaphongphat_game-based_2020} &
Pre/post test scores more than doubled \\
&
&
\cite{vathanakulkachorn_enhancing_2023} &
Test scores rose by 19.80\% post-intervention \\
&
&
\cite{wang_codingprogramming_2019,huang_design_2017} &
Significant learning gains and improvement in motivation post-intervention for robotics skills \\
\hline
Gamified &
Motivational and quantitative learning gains &
\cite{atmatzidou_didactical_2017} &
Sixth-graders’ computational-thinking rubric scores significantly rose from 2.10 to 2.72 out of 4.00 ($p < .001$) post-intervention \\
&
&
\cite{plaza_scratch_2019,plaza_stem_2019} &
Universal post-test improvements and surges in self-declared curiosity \\
&
&
\cite{eliza_game-d_2025} &
Statistically significant growth in algorithmic knowledge \\
&
&
\cite{canas_open-source_2020} &
Users rated the practice environment ``excellent'' (57.00\%) or ``satisfactory'' (43.00\%) \\
&
&
\cite{praveena_effective_2024} &
Average course scores above 83.00\% alongside high engagement and motivation indices (\textgreater{} 4.5 out of 5.0) \\
&
&
\cite{lin_applying_2022} &
Average course scores above 83.00\% alongside high engagement and motivation indices (\textgreater{} 80.00\%) \\
\hline
\end{tabular}
\end{table}

\subsubsection{Comparative Effectiveness of Formal and Informal Learning Contexts.}\label{results_outcomes_context}

The summary of learning outcomes of formal and informal learning contexts across the reviewed studies is shown in Table~\ref{tab:outcome_context}. Formal contexts tended to report quantifiable learning gains more often than informal contexts (36.84\%, 95\% CI [26.63\%, 50.88\%] vs. 21.05\%, 95\% CI [11.13\%, 36.36\%]), with $\chi^2(1) = 2.68$, $p = .101$, Cramér's $V = 0.13$, OR $ = 2.19$ (95\% CI [0.85, 5.64]), and RR $ = 1.75$ (95\% CI [0.87, 3.54]). Affective outcomes were common in both contexts (formal 59.65\%, 95\% CI [47.52\%, 71.93\%]; informal 50.00\%, 95\% CI [34.81\%, 65.25\%]), with $\chi^2(1) = 0.86$, $p = .354$, Cramér's $V \approx 0.00$, OR $ = 1.48$ (95\% CI [0.65, 3.38]), and RR $ = 1.19$ (95\% CI [0.81, 1.75]). See the \href{https://osf.io/gq3a8/files/4y8hn}{Supplementary Materials} for raw counts.

\begin{table}[htbp]
\centering
\footnotesize
\caption{The summary of learning outcomes of formal and informal learning contexts across the reviewed studies.}
\label{tab:outcome_context}

\renewcommand{\arraystretch}{1.2}
\begin{tabular}{p{0.15\linewidth} p{0.22\linewidth} p{0.12\linewidth} p{0.35\linewidth}}
\hline
\textbf{Learning context} &
\textbf{Positive outcome cluster} &
\textbf{References} &
\textbf{Detailed outcomes} \\
\hline
Formal &
Quantifiable learning gains &
\cite{duraes_gaming_2015} &
Significant score improvements (20--40\%) in a middle-school program that blended gaming challenges with LEGO robotics~ \\
&
&
\cite{atmatzidou_didactical_2017} &
Large pre-post jumps in computational-thinking tests~ \\
&
&
\cite{crespo_virtual_2015,crespo_virtual_2015-1} &
Students completed industrial-robot tasks more successfully when supported by a VR programming sandbox and judged a parallel simulator ``highly effective'' for mastering coordinate frames~ \\
&
&
\cite{leonard_using_2016} &
Boosted self-efficacy~ \\
&
&
\cite{thanyaphongphat_game-based_2020} &
Significant knowledge gains in robotics programming~ \\
&
&
\cite{kobayashi_design_2018} &
68.00\% of freshman teams completed at least one embedded-systems task in a robotics competition game \\
&
&
\cite{quintero_gamification_2022} &
Most participants completed all tasks in a gamified augmented educational application \\
&
&
\cite{watanabe_compulsory_2018} &
Every team mastered at least one mid-tier challenge in a robot contest \\
\hline
Informal &
Affective and accessibility outcomes &
\cite{yadagiri_blocks-based_2015} &
95.00\% of K-12 learners felt ``comfortable'' coding robots, and 90.00\% reported conceptual understanding \\
&
&
\cite{giang_exploring_2020} &
\textgreater{} 89.00\% self-reported engagement and enjoyment in an escape game-based robotics education scenario \\
&
&
\cite{heljakka_gamified_2019} &
Increased preschoolers' motivation to learn coding in a gamified toy robot coding study~ \\
&
&
\cite{yamazaki_comparative_2015} &
Raised motivation and interest by factors of 1.28 and 1.84, respectively \\
&
&
\cite{shim_effects_2016} &
Ease-of-use and enjoyment exceeded 4.2 out of 5.0, and confidence rose from 4.4 to 5.0 post-intervention \\
\hline
\end{tabular}
\end{table}


\subsubsection{Comparative Effectiveness of Pedagogical Models.}\label{results_outcomes_pedagogy}

Among the three pedagogical models observed across the reviewed studies, constructivist interventions predominantly targeted cognitive outcomes, experiential designs consistently boosted learners' motivation, confidence, and short-term task performance, and PBL fostered collaboration, self-efficacy, and authentic problem-solving (see Table~\ref{tab:outcome_pedagogy}).

\begin{table}[htbp]
\centering
\footnotesize
\caption{The summary of learning outcomes for the three types of pedagogical models used across the reviewed studies.}
\label{tab:outcome_pedagogy}

\renewcommand{\arraystretch}{1.2}
\begin{tabular}{p{0.15\linewidth} p{0.22\linewidth} p{0.12\linewidth} p{0.35\linewidth}}
\hline
\textbf{Pedagogical model} &
\textbf{Positive outcome cluster} &
\textbf{References} &
\textbf{Detailed outcomes} \\
\hline
Constructivist &
Cognitive gains &
\cite{eliza_game-d_2025} &
Enhanced learners’ conceptual understanding of straight-line motion \\
&
&
\cite{budiman_gamification_2025} &
Fostered critical thinking and problem-solving skills \\
&
&
\cite{lee_learning_2019} &
Widespread agreement on the successful development of computational thinking skills \\
&
&
\cite{moreno-vera_comparison_2018} &
Facilitated the transfer of abstract logic skills across diverse age groups \\
\hline
Experiential &
Increased motivation, confidence, and short-term task performance &
\cite{kunovic_comparison_2021} &
High proportions of students reported enjoyment and perceived usefulness. High self-reported interest and motivation (\textgreater{} 4 out of 5; \textgreater{} 80\%) in robotics and related fields \\
&
&
\cite{budiman_gamification_2025} &
Hands-on competence gains. High self-reported interest and motivation (\textgreater{} 4 out of 5; \textgreater{} 80\%) in robotics and related fields \\
\hline
Project-based~ &
Collaboration, self-efficacy, and authentic problem-solving~ &
\cite{panwar_analyzing_2020,sarkar_teaching_2020} &
Teams were able to refine complex robot designs iteratively \\
&
&
\cite{kashinath_narvekar_learn_2020,lin_applying_2022} &
Marked improvements in domain-specific skills \\
&
&
\cite{leonard_using_2016,plaza_scratch_2019,plaza_stem_2019} &
Positive shifts in STEM attitudes \\
\hline
\end{tabular}
\end{table}

\subsubsection{Analysis of Learning Outcomes of Different Levels of Robotics-related Skills.}\label{results_outcomes_skills}

The effects of different levels of robotics-related skills on learning outcomes were diverse (see Table~\ref{tab:outcome_skills}). Basic-level interventions, such as block-based sequencing activities, small script segments, and basic sensor integration and electronic circuitry, primarily emphasized affective benefits. Intermediate-level courses, where learners engaged in tasks such as writing multi-file programs, manipulating libraries, integrating sensors, and multi-component integration, similarly reported motivational gains as well as learning gains. Lastly, advanced-level programs also consistently demonstrated efficiency and usability gains.

\begin{table}[htbp]
\centering
\footnotesize
\caption{The summary of learning outcomes for the three levels of programming/software development and understanding of robotic systems skills across the reviewed studies.}
\label{tab:outcome_skills}

\renewcommand{\arraystretch}{1.2}
\begin{tabular}{p{0.15\linewidth} p{0.22\linewidth} p{0.12\linewidth} p{0.35\linewidth}}
\hline
\textbf{Skill level} &
\textbf{Positive outcome cluster} &
\textbf{References} &
\textbf{Detailed outcomes} \\
\hline
Basic &
Affective benefits &
\cite{leonard_using_2016} &
Children demonstrated higher STEM self-efficacy than control groups~ \\
&
&
\cite{dahal_introductory_2023} &
Novice learners successfully wired \textit{SmartMotors} in introductory robotics activities,~ \\
&
&
\cite{kunovic_comparison_2021,quintero_gamification_2022} &
Reduced anxiety and greater inclusive participation~ \\
&
&
\cite{giang_exploring_2020} &
Enjoyment and flow ratings exceeding 90.00\% in an escape-room scenario~ \\
&
&
\cite{praveena_effective_2024} &
Motivation and engagement scores between 80.00--90.00\% for a VR-based assembly module in a first-year engineering course \\
&
&
\cite{panskyi_holistic_2021} &
Primary school pupils reported high self-rated creativity (4.70/5.00) and interest in robotics (4.74/5.00)~ \\
&
&
\cite{lee_learning_2019} &
Enabled 70.00--90.00\% of secondary students to master recursion and problem decomposition \\
&
&
\cite{yamazaki_comparative_2015} &
Nearly doubled motivation and interest in robotics \\
&
&
\cite{sharma_coding_2019} &
High levels of enjoyment (4.6/5.0) and post-activity coding interest (4.4/5.0)~ \\
\hline
Intermediate &
Motivational and learning gains &
\cite{budiman_gamification_2025} &
92.00\% engagement and 90.00\% motivation in a line-following robot course \\
&
&
\cite{chung_design_2021} &
8.75\% increase in interest in STEM careers following an online robotics competition \\
&
&
\cite{han_skynetz_2016} &
Over 90.00\% of participants reported perceived knowledge gains~ \\
&
&
\cite{canas_open-source_2020} &
57.10\% of users rated the practice environment as ``excellent''~ \\
&
&
\cite{watanabe_compulsory_2018} &
100.00\% mid-level assignment completion rates in a game-based robotics contest~ \\
&
&
\cite{jadhav_pbl_2021} &
Fostered substantial learning and an increase in motivation \\
\hline
Advanced &
Efficiency and usability gains &
\cite{vazquez-hurtado_virtual_2024} &
Users reported maximum ease of learning scores (5 out of 5) in a smart-factory digital twin environment~ \\
&
&
\cite{costa_robotics_2016,crespo_virtual_2015,crespo_virtual_2015-1} &
100.00\% of participants reported knowledge acquisition~ \\
&
&
\cite{buckley_interdisciplinary_2022} &
76.00\% of high school students successfully explained neuroswarm behavior following a multidisciplinary intervention~ \\
&
&
\cite{peterson_evaluating_2022} &
4.30\% increase in motivation and a 1.90\% increase in learning effectiveness compared to traditional instruction \\
\hline
\end{tabular}
\end{table}

\subsubsection{Comparative Effectiveness of Intervention Types.}\label{results_outcomes_intervention}

Embodied, block-based workflows involving physical robots consistently demonstrated significant increases in self-reported interest, confidence, and motivation (see Table~\ref{tab:outcome_interv}). Simulator-centric approaches, often incorporating VR or distance-learning toolkits, increasingly emphasize accessibility gains and cognitive improvements (see Table~\ref{tab:outcome_interv}). Stand-alone digital games also showed notable learning gains and modest affective benefits (see Table~\ref{tab:outcome_interv}). Lastly, competitions and MOOCs in robotics education emphasized authentic problem-solving (see Table~\ref{tab:outcome_interv}).

\begin{table}[htbp]
\centering
\footnotesize
\caption{The summary of learning outcomes for different types of intervention methodologies utilized across the reviewed studies.}
\label{tab:outcome_interv}

\renewcommand{\arraystretch}{1.2}
\begin{tabular}{p{0.15\linewidth} p{0.22\linewidth} p{0.12\linewidth} p{0.35\linewidth}}
\hline
\textbf{Intervention type} &
\textbf{Positive outcome cluster} &
\textbf{References} &
\textbf{Detailed outcomes} \\
\hline
Block-based programming with physical robots &
Motivational and affective gains &
\cite{leonard_using_2016} &
Enhanced children's self-efficacy compared to control groups \\
&
&
\cite{ono_measuring_2024} &
Enjoyment rates exceeding 96.00\% \\
&
&
\cite{nordin_mobile_2020,zainal_primary_2018} &
Self-reported scores above 3.9 out of 5 for learning, motivation, and interest \\
&
&
\cite{yadagiri_blocks-based_2015} &
95.00\% of participants expressed comfort in robot programming post-intervention \\
&
&
\cite{wang_codingprogramming_2019} &
Significant programming and robot design knowledge gains \\
&
&
\cite{kilic_exploring_2022} &
Acceptance, usefulness, and ease-of-use ratings above 4.4 out of 5.0 \\
&
&
\cite{giang_exploring_2020} &
100.00\% enjoyment with 84.00\% of learners intending to continue studying robotics \\
\hline
Simulators &
Accessibility gains and cognitive improvements &
\cite{crespo_virtual_2015,crespo_virtual_2015-1} &
Unanimous agreement on achieving learning goals. Ease-of-use ratings above 80.00\% \\
&
&
\cite{lucan_simulator-based_2023} &
Grade parity with traditional instruction \\
&
&
\cite{peterson_evaluating_2022} &
Significant post-test gains in confidence, motivation, and interest \\
&
&
\cite{praveena_effective_2024} &
Large effect sizes \\
&
&
\cite{costa_robotics_2016} &
100.00\% perceived effectiveness \\
&
&
\cite{canas_ros-based_2020} &
95.00\% positive evaluations \\
\hline
Digital games &
Notable learning gains &
\cite{alfieri_case_2015} &
Test scores rose from 6.74 to 7.97 out of 17 post-intervention \\
&
&
\cite{agalbato_robo_2018} &
\textgreater{}95.00\% task completion \\
&
&
\cite{matsumoto_relationship_2018} &
In-game success aligned with exam performance \\
&
Modest affective benefits &
\cite{gabriele_educational_2017} &
Slight increases in self-determination and self-efficacy \\
&
&
\cite{sochol_exploration_2024} &
Heightened enthusiasm for STEM design \\
\hline
Competitions and MOOCs~ &
Authentic problem-solving~ &
\cite{leung_project_2022} &
Broad educational content coverage \\
&
&
\cite{chung_design_2021} &
Superior user performance compared to control groups \\
&
&
\cite{lin_applying_2022,liu_stem_2014} &
Heightened interest and motivation in robotics and STEM fields \\
&
&
\cite{canas_open-source_2020,canas_ros-based_2020,roldan-alvarez_unibotics_2024} &
Enhanced user experiences relative to traditional instructional methods \\
&
&
\cite{jadhav_pbl_2021,lin_applying_2022} &
Substantial learning gains \\
&
&
\cite{vazquez-hurtado_adapting_2022} &
Acquisition of conceptual skills \\
&
&
\cite{maurelli_blackpearl_2023} &
Advantages over purely simulation-based environments \\
\hline
\end{tabular}
\end{table}

\subsubsection{Effect of Immersive Technology on Learning Outcomes.}\label{results_outcomes_immersive}

Across studies that incorporated immersive technologies, a consistent pattern of positive outcomes emerged, particularly around heightened engagement, improved spatial reasoning, and enhanced perceived realism. In a first-year robotic assembly course, head-mounted VR provided novices with ``hands-on'' assembly experiences rated as effective by 85.00\% of students, engaging by 90.00\%, and motivating by 80.00\%~\cite{praveena_effective_2024}. Comparable or superior outcomes relative to face-to-face laboratories were also reported in industrial-robot training studies, where VR groups completed tasks with fewer trajectory errors~\cite{peterson_evaluating_2022} than desktop users, while an augmented reality (AR)-enhanced mBot course achieved significantly higher skill-completion rates than a non-AR control group~\cite{quintero_gamification_2022}. Even lightweight implementations, such as short VR puzzle segments, generated unanimous enthusiasm and enjoyment~\cite{karageorgiou_escape_2019}. Similarly, another VR-based robotics simulator study highlighted VR's primary instructional contribution: vivid, first-person visualization that reduced the cognitive burden associated with mastering spatially complex concepts~\cite{crespo_virtual_2015-1}. 

Only five empirical investigations have incorporated haptic feedback into game-based and gamified robotics education, and they collectively suggest that tactile cues can deepen immersion. Three studies reporting on the \textit{Robotic Academy} VR platform~\cite{peterson_evaluating_2022,peterson_teaching_2021,vassigh_performance-driven_2023} showed that adding force-feedback to industrial-robot simulators yielded modest but measurable performance gains: hands-on test scores rose 1.90\% over conventional training, robot-anatomy errors fell by 53.34\%, and self-reported confidence, motivation, and interest exceeded classroom baselines by roughly 3-4\%. Haptics also proved invaluable for accessibility: a Wiimote-based vibration channel enabled students with visual impairments to follow robot movements, with 97.00\% reporting feeling capable of working with computers or robots after the activity~\cite{park_engaging_2013}. Similarly, the ARSIS 6.0 mixed-reality toolkit described perceived benefits of its haptic tool affordances, such as intuitive interaction, ergonomic consistency, and immersive feedback~\cite{teogalbo_mixed_2023}.

\subsection{Identified Challenges and Limitations}\label{results_challenges}

Although this systematic review of the literature uncovered valuable insights into how GBL and gamification influence educational outcomes in robotics, several challenges were also identified. The following section explores the key limitations and challenges associated with these methods, providing a foundation for addressing \textbf{RQ3}: What are the limitations and challenges of game-based and gamified approaches to teaching robotics and associated skills?

\subsubsection{Challenges of Different Approach Types}\label{results_challenges_approach}

GBL studies most frequently cited resource intensity, usage fatigue, and evaluation gaps: high development and production costs~\cite{hrbacek_step_2018}, VR-sickness or simulator fatigue~\cite{crespo_virtual_2015-1}, and the lack of detailed assessment and statistical tests and/or usage of small samples~\cite{dahal_introductory_2023,duraes_gaming_2015,merkouris_programming_2021,thanyaphongphat_game-based_2020,yuen_mobile_2021}. Drop-out spikes (12.50\% in Chung et al.~\cite{chung_design_2021} and 18.00\% in Leonard et al.~\cite{leonard_using_2016}) further signified the challenge of sustaining engagement once novelty wore off. Gamified studies, while generally cheaper to deploy, often reported cognitive overload and competitive stress: ROS set-up complexity deterred novices in a drone programming course~\cite{canas_open-source_2020}, students needed ``significant instructor support'' in Plaza et al.~\cite{plaza_scratch_2019}, and one learner felt ``extremely demotivated'' by the high-stakes competitive format in Mart\'in~\cite{martin_open_2018}. Technical glitches (e.g., software crashes in a line-follower task~\cite{eliza_game-d_2025}) and low completion rates (13.30\%) in an aquatic robot-fish challenge~\cite{kashinath_narvekar_learn_2020} were also recurrent.

\subsubsection{Challenges of Different Learning Contexts}\label{results_challenges_context}

Common shortcomings were observed across both learning contexts. Several formal studies relied on self-report or omitted statistical tests despite substantial learning gains~\cite{crespo_virtual_2015-1,duraes_gaming_2015}. Others flagged resource or difficulty constraints: a college LEGO-League module required high time commitment and advantaged prior-experts~\cite{strnad_programming_2017}; only 46.00\% of freshmen teams completed an electronics task in a PBL course~\cite{kobayashi_design_2018}; and 60.00\% of teams were unable to complete a single challenge in a robotics competition~\cite{sarkar_teaching_2020}. Informal projects were found to be even more likely to prioritize exposure over measurement: first-graders introduced to loops through storytelling games~\cite{sovic_how_2014}, an online engineering mini-game~\cite{jansen_pump_2016}, and a build-and-play robotics club~\cite{roscoe_teaching_2014} all reported no formal learning data. A visionary but hardware-intensive remote-learning framework likewise omitted outcome evidence~\cite{hempe_erobotics_2015}. Where quantitative data were gathered, effects were sometimes modest or non-significant. For example, the motivational gain in a comparative study on programmable
robots as programming educational tools did not reach statistical significance~\cite{yamazaki_comparative_2015}.

\subsubsection{Challenges of Different Pedagogical Models}\label{results_challenges_pedagogy}

Several negative outcomes were observed across all three pedagogical models. For example, in constructivist interventions, initial cognitive gains often diminished rapidly over time~\cite{moreno-vera_comparison_2018}, and quiz-based assessments occasionally revealed lower knowledge retention compared to traditional instruction~\cite{ono_measuring_2024}. Experiential designs reported technical and logistical limitations: small samples~\cite{ates_work_2022,crespo_virtual_2015-1,han_skynetz_2016,mendonca_digital_2020,yadagiri_blocks-based_2015}, frustration with hardware inconsistencies and software bugs~\cite{kunovic_comparison_2021,praveena_effective_2024}, and reduced tactile interaction in purely virtual settings~\cite{nordin_mobile_2020,praveena_effective_2024}. Lastly, the reported trade-off for PBL, echoed across Panwar et al.~\cite{panwar_analyzing_2020}, Plaza et al.~\cite{plaza_scratch_2019}, Lin et al.~\cite{lin_applying_2022}, and Kobayashi et al.~\cite{kobayashi_design_2018}, was the heavy time, resource, and instructor demand of sustained projects, leading to partial task completion and unequal contribution levels.

\subsubsection{Challenges across Different Levels of Robotics-related Skills}\label{results_challenges_skills}

For basic-level interventions, software glitches and the absence of physical interaction opportunities were reported to dampen learner enthusiasm~\cite{praveena_effective_2024}, while holistic programs imposed substantial resource demands~\cite{panskyi_holistic_2021}. Some activities, such as escape-room tasks~\cite{giang_exploring_2020}, proved insufficiently challenging for experienced learners. Furthermore, certain studies, including a bio-inspired prototyping kit~\cite{bas_enhancing_2023} and co-robotic games~\cite{higashi_design_2021}, reported no outcome data at all. Additionally, reliance on pre-built robots~\cite{yadagiri_blocks-based_2015} and pre-programmed elements~\cite{kunovic_comparison_2021} risked fostering only a superficial understanding of robot hardware and software, with hardware inconsistencies occasionally causing learner frustration~\cite{kunovic_comparison_2021}.

Several intermediate-level studies, including Sarkar et al.~\cite{sarkar_teaching_2022}, reported minimal outcome analytics and steep technical barriers, such as the complexity of ROS in a drone programming course~\cite{canas_open-source_2020}, posing significant hurdles for novices. User experience ratings were also mixed, with only 18.00\% positive evaluations for \textit{Skynetz}~\cite{han_skynetz_2016}. Additionally, completion fatigue became apparent as task difficulty increased, evidenced by a 0.00\% success rate on an AR-marker task in an embedded-systems contest~\cite{watanabe_compulsory_2018} and a 12.45\% dropout rate in an online competition~\cite{chung_design_2021}. The substantial time demands noted in another online competition~\cite{panwar_analyzing_2020} further restricted sustained participation.

In advanced-level studies, authors noted novelty effects that occasionally diminished post-quiz retention despite high engagement~\cite{peterson_evaluating_2022}, and steep learning curves~\cite{costa_robotics_2016,gabriele_educational_2017,hempe_erobotics_2015}, with one-third of participants unable to complete tasks in a ROS-based module~\cite{costa_robotics_2016}. Additionally, persistent ergonomic and technical challenges, such as software compatibility issues~\cite{buckley_interdisciplinary_2022} and user discomfort during prolonged use~\cite{osborne_wip_2024} were consistently reported across these studies.

\subsubsection{Challenges of Different Intervention Types}\label{results_challenges_intervention}

Studies utilizing block-based programming with physical robots reported 18.50\% dropout rate~\cite{leonard_using_2016}, reduced opportunities for hands-on manipulation~\cite{nordin_mobile_2020}, lower quiz performance despite high enjoyment levels~\cite{ono_measuring_2024}, and moderate challenge levels causing boredom among more advanced learners~\cite{zainal_primary_2018}. Simulator-based approaches suffered from steep technical learning curves and technical complexity, which impeded task completion in ROS-based simulators~\cite{costa_robotics_2016}, and lower written quiz scores compared to traditional instruction~\cite{peterson_evaluating_2022}. For stand-alone digital games, authors noted negative outcomes such as non-significant motivational changes~\cite{gabriele_educational_2017}, frustration with tutorials~\cite{alfieri_case_2015}, constrained coding practice due to oversimplified instructions~\cite{agalbato_robo_2018}, competence gaps~\cite{matsumoto_relationship_2018}, and high dropout rates~\cite{sochol_exploration_2024}. Lastly, high resource and time investment requirements~\cite{lin_applying_2022,panwar_analyzing_2020,strnad_programming_2017}, notable attrition rates~\cite{chung_design_2021,kashinath_narvekar_learn_2020}, and anxiety-related declines in learner confidence~\cite{martin_open_2018} were reported as limitations in competition and MOOC-focused approaches.

\subsubsection{Challenges of Immersive Technology}\label{results_challenges_immersive}

Many VR studies relied on small or convenience samples or lacked detailed outcome assessments~\cite{asut_developing_2024,crespo_virtual_2015-1,hamann_gamification_2018,hempe_erobotics_2015,jansen_pump_2016,osborne_wip_2024,vazquez-hurtado_virtual_2024}, making it difficult to attribute learning gains solely to immersion. Additional challenges included reduced opportunities for physical interactions~\cite{hamann_gamification_2018,nordin_mobile_2020}, high equipment costs~\cite{hempe_erobotics_2015}, substantial implementation demands~\cite{hempe_erobotics_2015}, motion-related discomfort~\cite{osborne_wip_2024,praveena_effective_2024}, and the complexity of bridging virtual simulation models with real-world hardware applications~\cite{hamann_gamification_2018}.

The modest but measurable performance gains obtained by adding force-feedback to industrial-robot simulators on the \textit{Robotic Academy} VR platform~\cite{peterson_evaluating_2022,peterson_teaching_2021,vassigh_performance-driven_2023} were offset by slower written-quiz completion time (i.e., 43.00\% longer) and slightly lower written test results (i.e., 2.38\% lower), suggesting that the additional sensory load may impede the retrieval of declarative knowledge. In addition, the ARSIS 6.0 mixed-reality toolkit, which contained haptic tool affordances, provided no outcome data and flagged potential physical discomfort during extended sessions~\cite{teogalbo_mixed_2023}.

\subsection{Risk of Bias}

\subsubsection{MMAT Appraisal Summary}\label{results_rob_MMAT}

Of the 95 studies included in the review, 
most were quantitative descriptive (\(n = 61; 64.21\%\)), followed by quantitative non-randomized designs (\(n = 19; 20.00\%\)). Mixed-methods studies were rare (\(n = 3; 3.16\%\)), and only one study was qualitative (\(1.05\%\)). Eleven studies (\(11.58\%\)) failed the MMAT screening questions and therefore could not be fully appraised, but they remained included in the systematic review.

The appraisal revealed a strong dominance of descriptive designs, which limited causal inference and generalizability. The scarcity of mixed-methods studies constrained triangulation of qualitative and quantitative insights. Frequent reliance on self-report and the absence of standardized measures further restricted the strength of evidence.

Item‑level profiles emphasize that descriptive reports often lack sampling/representativeness detail and non‑response analyses, even when measurements are adequate. In addition, non-randomized studies rarely describe intervention fidelity clearly despite strong measurement and complete outcomes. Lastly, mixed‑methods designs are under‑represented and typically under‑report integration and strand‑level quality. Full appraisal details are provided in the \href{https://osf.io/gq3a8/files/rz8h7}{Supplementary Materials} and in the \href{https://osf.io/gq3a8/files/tgamw}{Supplementary Table}.



\subsubsection{ROBINS-I Summary}\label{results_rob_robins}

We completed ROBINS-I appraisals for 19 quantitative non-randomized studies in our corpus. Overall risk of bias was ``Serious'' in 16/19 (84.21\%) and ``Moderate'' in 3/19 (15.79\%). No studies were judged ``Critical'' or ``Low'' overall.


Confounding (D1) was the principal concern (15 ``Serious,'' 4 ``Moderate''), reflecting the common absence of pre-specified control groups, baseline equivalence checks, or statistical adjustment for key covariates. Selection of participants (D2) also frequently raised concerns (15 ``Moderate,'' 4 ``Serious''), often due to convenience sampling or voluntary participation in clubs/competitions.

In contrast, classification of interventions (D3) was usually clear (17 ``Low,'' 2 ``Moderate''), as were deviations from intended interventions (D4; 13 ``Low,'' 6 ``Moderate''). Missing data (D5), measurement of outcomes (D6), and selective reporting (D7) were mostly ``Low'' to ``Moderate'' with one ``Serious'' instance each. Details are provided in the \href{https://osf.io/gq3a8/files/rz8h7}{Supplementary Materials} and in the \href{https://osf.io/gq3a8/files/sh6xf}{Supplementary Table}.


\section{DISCUSSION}\label{discussion}
\subsection{General Overview}\label{discussion_overview}

This systematic literature review synthesized findings from 95 studies on GBL and gamified approaches to robotics education, offering both broad insights and fine-grained thematic analysis. Extending beyond prior systematic reviews and meta-analyses that treated GBL and gamified STEM learning~\cite{ouyang_effects_2024,sailer_gamification_2020} and robotics education~\cite{angel-fernandez_towards_2018,sailer_gamification_2020,scaradozzi_towards_2019} as homogeneous domains, the following sections disaggregate findings by instructional approach, learning context, pedagogy, skill emphasis, and interface modality. Additionally, we propose a structured design space, detailing recommended practices, pitfalls to avoid, and directions for future research.


\subsection{Approach–Context–Pedagogy: The Design Space}\label{discussion_coupling}

Across settings, we observed a stable approach–context coupling: lightweight gamification concentrates in formal, credit-bearing courses that demand tight curricular alignment and predictable assessment, whereas fully fledged GBL appears more evenly distributed and frequently in informal or extracurricular contexts where schedules and evaluation are flexible (see Figure~\ref{fig:approach_x_context}). The pattern is not deterministic but reflects integration costs and affordances: gamified elements (e.g., points, badges, leaderboards) can be layered onto existing routines with minimal disruption, while GBL typically requires specialized content, longer sessions, and alignment with constructivist or experiential pedagogies. 



Our inferential analysis found no reliable difference between GBL and gamified approaches in the likelihood of reporting learning gains. Similarly, affective outcomes were broadly comparable across GBL and gamified studies. Formal contexts were more likely to report learning gains but faced workload and scalability issues, while informal settings enhanced enthusiasm and accessibility yet struggled to substantiate durable learning outcomes. These findings cohere with self-determination theory~\cite{ryan_self-determination_2000} (SDT), where autonomy, competence, and relatedness foster intrinsic motivation and persistence in learning tasks \cite{birk2016motivational}. In SDT terms, differences between GBL and gamification can be interpreted more precisely. Gamified systems in formal curricula typically emphasize structured tasks, progress cues, and feedback loops that foreground competence and externally recognized progress. In contrast, GBL implementations in informal or extracurricular settings (e.g., clubs, camps, and competitions) more often highlight open-ended challenges and collaborative problem-solving that support autonomy and relatedness. SDT thus helps clarify how similar surface mechanics (e.g., points, badges, levels, challenges) can lead to different motivational patterns depending on how they are configured to support (or frustrate) these three basic psychological needs in context.

Pedagogy mediates both design and outcomes. Informal GBL interventions frequently embed experiential or constructivist cycles (problem encounter, artifact creation, reflection), supporting systems-level reasoning and iterative debugging, whereas classroom gamification more often scaffolds micro-behaviors (participation, practice cadence, timely submission) aligned with formal course flow and assessment structures, sometimes via explicit project management scaffolds. This pattern aligns with our statistical analysis showing that constructivist pedagogies appeared more often in GBL, while project-based pedagogies were more common in gamified studies.

Two implications follow. First, transposition is possible in both directions: modular, time-bounded GBL episodes can fit formal courses when anchored to explicit objectives and rubrics, and informal programs can exploit low-cost gamification to sustain momentum between sessions. Second, study design should foreground pedagogy as the mediating lever between approach and context rather than treating those as independent variables.

\subsection{Skills and Progression: From Novice Onboarding to Advanced Competence}\label{discussion_skills}

Skill mapping reveals a strong emphasis on introductory programming and modular kit assembly, with comparatively fewer interventions addressing advanced software development (e.g., ROS node development~\cite{costa_robotics_2016}) or integrated systems work~\cite{hamann_gamification_2018}. A small subset tackled higher-level robotic systems skills, while several advanced efforts reported limited or no tangible outcomes, highlighting persistent evaluation gaps~\cite{hamann_gamification_2018,hempe_erobotics_2015}. At the entry level, many interventions report boosts to self-efficacy, motivation, and enjoyment (e.g.,~\cite{dahal_introductory_2023,panskyi_holistic_2021,quintero_gamification_2022,sharma_coding_2019}) but also note shallow hardware understanding or insufficient challenge for advanced learners~\cite{giang_exploring_2020,kunovic_comparison_2021,praveena_effective_2024}. Descriptively, intermediate offerings show skill gains~\cite{budiman_gamification_2025,canas_open-source_2020,han_skynetz_2016}, but encountered steep technical barriers and significant dropout rates~\cite{chung_design_2021,sarkar_teaching_2022,watanabe_compulsory_2018}. Advanced programs demonstrate promising outcomes~\cite{peterson_evaluating_2022,vazquez-hurtado_virtual_2024} but were constrained by small samples, novelty effects, and technical hurdles~\cite{buckley_interdisciplinary_2022,costa_robotics_2016,gabriele_educational_2017,osborne_wip_2024}. Our inferential analysis showed that programming/software development skill tiers differed by approach, with gamified studies more often targeting higher levels, whereas robotics systems skill tiers did not differ.

To close these curricular gaps, we propose a staged progression model that aligns tasks, scaffolds, and assessments (see Figure~\ref{fig:design_progression}). In addition, as suggested by Altmeyer et al.~\cite{altmeyer2018investigating}, gamification mechanics should adapt to learners’ readiness stages, with elements such as badges, challenges, and social features becoming more persuasive at higher stages. Skill progression needs to be carefully designed across the novice-to-expert continuum, employing dynamic difficulty adjustment (DDA), personalized feedback, and scaffolded challenges that adapt to learner competence. Comparable work by Karaosmanoglu et al.~\cite{karaosmanoglu2024born} catalogs before-game and in-game adaptations, such as subjective-rating, performance-based, and physiological signal-based adaptations, offering a ready template for DDA in game-mediated robotics education. Moreover, as improvements typically follow a power law of practice, progression-aware designs should anticipate plateaus in perceived competence and employ staged novelty and scaffolding to sustain motivation, an effect echoed in a study by Birk et al.~\cite{birk2016motivational}. Assessments should deepen across stages: start with short, targeted tasks scored with simple rubrics and brief observer checklists for novices; add artifacts/process analytics later.
\vspace{1cm}


\begin{quote}
\textbf{Example 1: Game-based embedded systems robot contest (GBL, Formal)}

Watanabe et al.~\cite{watanabe_compulsory_2018} describe a compulsory game-based contest embedded in a multi-week embedded-systems curriculum. Student teams use a mobile robot with sensors and a camera, progressing through a sequence of mini-games: 1) spin turn (precise rotation control), 2) line trace (following a line smoothly using basic robot control algorithms), 3) maze (obstacle-avoiding navigation), 4) AR-marker (detecting and reporting AR marker positions), before a final auto-cruising challenge requiring integration of all prior skills. Performance scores and a technical poster presentation determine the final evaluation. \\

From an SDT perspective, the design foregrounds competence via clear performance criteria, while its multi-team format provides relatedness through collaboration and peer support.

\end{quote}

\subsection{Modalities, Genres, and Hybrid Pathways}\label{discussion_modalities}

Different modalities afford different kinds of learning. Physical robots foreground material constraints and embodied troubleshooting (e.g., LEGO Mindstorms~\cite{leonard_using_2016}, Lemon~\cite{bas_enhancing_2023}, SmartMotors~\cite{dahal_introductory_2023}); simulators (e.g., ROS~\cite{costa_robotics_2016}, Gazebo~\cite{canas_ros-based_2020}, Webots~\cite{hamann_gamification_2018}) offer scalable, safe practice with rich scenario control, though some comparisons show lower user performance or preference versus traditional / hardware-based instruction, pointing to cognitive-load and fidelity gaps that need careful management~\cite{peterson_evaluating_2022,hamann_gamification_2018}. Digital games (including commercial inspirations~\cite{sochol_exploration_2024} and education-specific titles~\cite{matsumoto_relationship_2018,agalbato_robo_2018}) excel at motivation and pacing but must be orchestrated to sustain depth (e.g., genre-tailored practice loops). Block-based interfaces (e.g., Scratch~\cite{plaza_scratch_2019} and Blockly~\cite{merkouris_programming_2021}) remain effective on-ramps with known limits on conceptual depth unless paired with stronger evaluative scaffolds. Competitive formats (competitions/MOOCs) offer scalability and authenticity but often lack rigorous measures of learning gains and face challenges related to complexity and participant attrition~\cite{canas_open-source_2020,fernandez-ruiz_automatic_2022,roldan-alvarez_unibotics_2024}.

Genre should be selected for its pedagogical affordance: puzzle/strategy mechanics scaffold discrete concepts and systematic debugging~\cite{kilic_exploring_2022,costa_robotics_2016}, while simulation/sandbox mechanics suit experiential systems thinking and iterative design~\cite{canas_open-source_2020,hamann_gamification_2018}. The distribution of genres and mechanics in our corpus also reflects different affordances for how learners engage with robotics tasks (e.g., \cite{norman2013design,gaver1991technology}). Puzzle-oriented games foreground short, discrete problem-solving episodes that align well with introductory syntax and logic exercises. Simulation and sandbox environments, by contrast, afford extended experimentation with system-level behaviors such as control loops, sensing, and multi-robot coordination. From an HCI perspective, choosing a game genre is therefore not just an aesthetic decision but a way of structuring which forms of robotics practice become easy, frequent, and rewarding for learners. In practice, many of the strongest designs combine different genres and/or modalities (e.g., \emph{Simulate $\rightarrow$ Build $\rightarrow$ Test $\rightarrow$ Reflect}) to balance cost, fidelity, and motivation. Evidence on transfer supports this hybrid logic: VR drill-and-practice simulations have matched or exceeded in-person training on manipulator skills and post-assembly tasks~\cite{peterson_teaching_2021,vassigh_performance-driven_2023,praveena_effective_2024,peterson_evaluating_2022}; perception-based metrics and self-reports on VR-based simulators also indicate high perceived usefulness and behavioral intention for future deployment~\cite{kilic_exploring_2022}. At the same time, limitations such as limited haptics, simulation–reality mismatches, and setup/comfort issues can dampen far transfer and persistence ~\cite{asut_developing_2024,nordin_mobile_2020,peterson_evaluating_2022,sochol_exploration_2024,teogalbo_mixed_2023}. Therefore, as learners move from virtual to tangible system, bridging tasks and later physical validation are crucial to help preserve structure and motivation~\cite{karageorgiou_escape_2019,hamann_gamification_2018}. From an embodiment and presence perspective, our findings on immersive modalities suggest that VR and haptics do more than simply add immersion~\cite{dede_immersive_2009,lee_learning_2014,ratan_self-presence_2013,shin_role_2017,steed_impact_2016}. They reconfigure how learners perceive and act on robot behaviors by coupling body movement, spatial perception, and multisensory feedback into the learning loop~\cite{mubarrat_evaluation_2020,mubarrat_physics-based_2024,mubarrat_evaluating_2024,srinivasan_biomechanical_2021}. These systems create embodied affordances that can support motivation, engagement, and performance, as reflected in XR exergames and social VR~\cite{karaosmanoglu2024born,freeman2021body,freeman2022working} and in haptic-enhanced training simulations that improve training outcomes~\cite{ harvey_comparison_2021,skola_progressive_2019,grant_audiohaptic_2019}.

\begin{quote}
\textbf{Example 2: AR for robot assembly and programming (Gamified, Formal)}

Quintero et al.~\cite{quintero_gamification_2022} developed AR-mBot, a gamified augmented reality application designed to teach secondary-school robotics. Learners progress through four levels: Introduction, Assembly, Connection, and Programming. Each level contains different lessons and a challenge. Gamification elements include trophies, ``patches'' (badges), experience points, level progression, and cosmetic rewards. Students worked in groups, with AR overlays that supported them in discovering robot parts during assembly tasks. Students using AR-mBot significantly outperformed a control group on a robotics knowledge test (medium effect size), and most progressed through all levels of the application. \\

This case demonstrates how mobile AR with simple gamification mechanics can make step-by-step robotics instruction more effective.

\end{quote}

\subsection{Equity and Emerging Technologies: Accessibility, VR/Haptics, and AI}\label{discussion_equity}

Across the studies reviewed, accessibility and inclusivity remain inconsistently addressed. Prior reviews note that accessible design principles are often under-specified~\cite{miljanovic_review_2018}, and our corpus showed that relatively few interventions explicitly targeted diverse learner groups. Notable exceptions include work with learners with visual impairments~\cite{park_engaging_2013}, gender-equity efforts~\cite{wang_codingprogramming_2019}, and programs in resource-constrained or culturally diverse settings~\cite{jadhav_pbl_2021,panwar_analyzing_2020,vathanakulkachorn_enhancing_2023,orozco_gomez_ubuingenio_2023}. Inclusive designs are critical, as parallel HCI work by Freeman et al.~\cite{freeman2021body} in social VR shows how avatar design, gender presentation, and ethnicity can shape attention, stigma, and harassment. Such designs typically share traits such as \emph{low-floor/high-ceiling} task structures, adjustable pacing, multimodal feedback, explicit wayfinding and error recovery, and collaborative mechanics that distribute expertise. Additionally, age inclusivity is essential, as a study by Altmeyer et al.~\cite{altmeyer2018investigating} reported that older adults (75+) prioritize social interaction and collaboration in gaming while showing little interest in points, badges, and leaderboards. Consequently, age-aware game-mediated robotics learning should minimize competitive elements, provide privacy options, and incorporate virtual companions to foster autonomy and relatedness. 

Immersive technologies (VR/haptics) are promising but underused relative to their potential for presence, embodiment, and safe exposure to rare or risky scenarios. When used, they should be engineered for comfort and access and evaluated beyond engagement (performance, retention, transfer) [11, 135, 141, 171] as issues of safety, adaptation, and hardware constraints have been found to limit scalability despite strong motivational benefits in a similar XR exergame research study by Karaosmanoglu et al.~\cite{karaosmanoglu2024born}. Beyond technical fidelity, future research should consider how embodied presence and identity dynamics (well-documented in social VR platforms \cite{freeman2021body,freeman2022working}) shape learner motivation and collaboration in robotics education. 

Artificial Intelligence (AI) integration is nascent: The \textit{Robotic Academy} VR platform's Adaptive Learning System~\cite{peterson_evaluating_2022,peterson_teaching_2021,vassigh_performance-driven_2023}, \textit{BrainFarm}’s evolutionary robotics~\cite{gabriele_educational_2017}, and \textit{SmartMotors}’ sensor-based ``training'' ~\cite{dahal_introductory_2023} exemplify early patterns, alongside computer-vision applications in competitions and domain-specific tasks~\cite{leung_project_2022,lin_applying_2022,jadhav_pbl_2021}. Near-term opportunities include competency-aware coaching (detecting wiring/logic/parameter failure modes) and generative practice aligned to specific objectives, balanced with transparency and validity concerns.

AI, inclusivity, and immersive technologies appear only as emerging considerations within this review, reflecting their limited representation in the corpus and their role as future research directions rather than primary contribution threads.

\begin{quote}
\textbf{Example 3: Multimodal robotics for visually impaired learners (GBL, Informal)}

Park and Howard \cite{park_engaging_2013} describe a game-based robotics activity designed for middle and high school students with visual impairments. Learners programmed and played with mobile robots in a series of games, using a Wiimote configured as a vibrotactile channel with distinct vibration patterns signaling ``travel forward,'' ``turn left,'' ``turn right,'' ``object detected,'' ``bump,'' and ``goal achieved.'' The multimodal feedback, combining haptic and audio cues, allowed students to follow the robot's behavior without relying on visual inspection. Outcomes were strongly positive, with 97\% reporting feeling capable of working with computers or robots after the camp. \\

This case is a concrete example of how haptic-audio multimodal feedback can make robotics experiences more accessible and nonvisually navigable for learners with visual impairments.

\end{quote}

\subsection{Implications for Designers and Recommendations for Future Research}\label{discussion_designSpace}

\begin{figure*}[hbt]
\centering
\begin{subfigure}{\linewidth}
    \centering
    \includegraphics[width=1.0\linewidth]{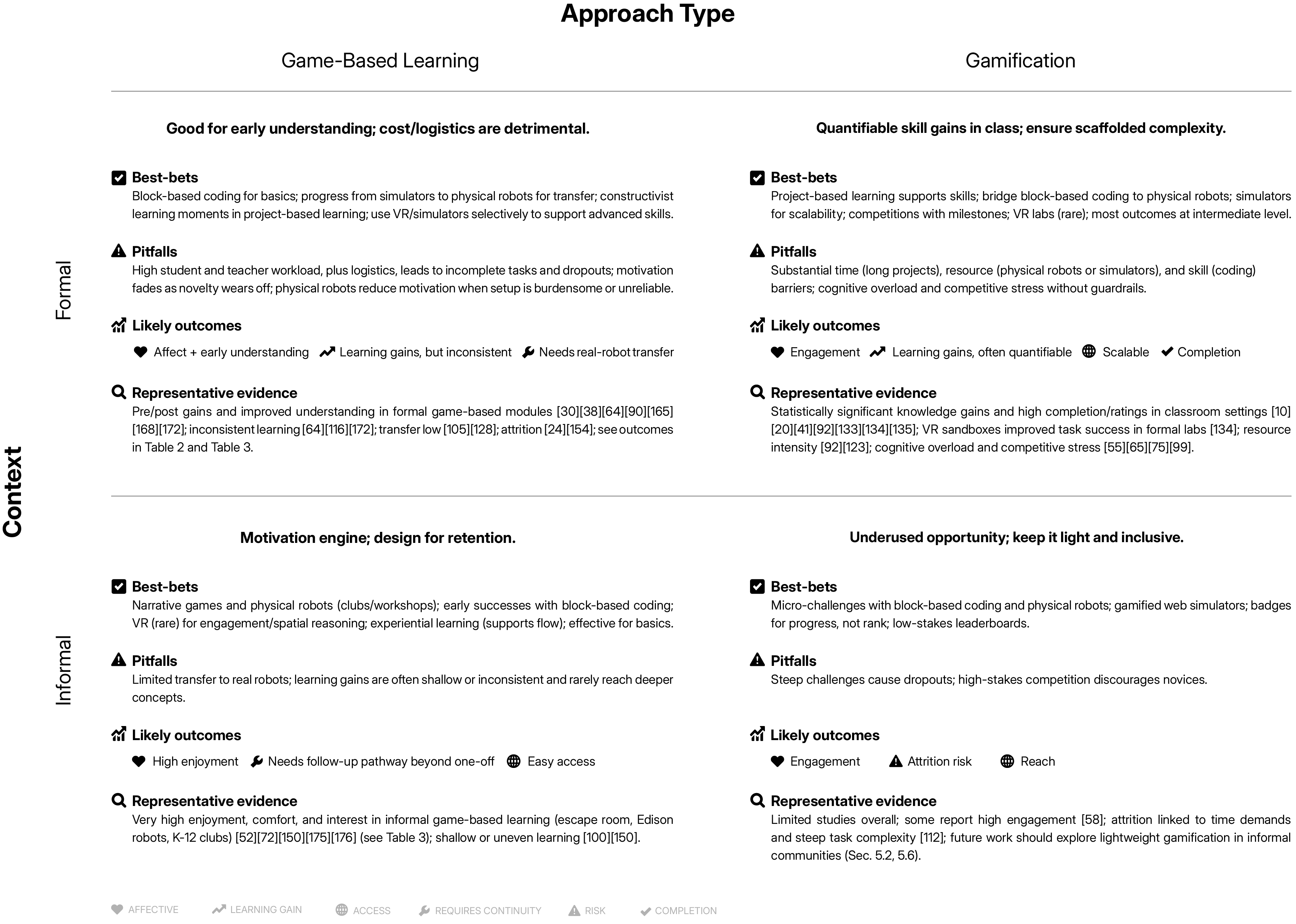}
\end{subfigure}

\vspace{1em} 

\begin{subfigure}{\linewidth}
    \centering
    \includegraphics[width=1.0\linewidth]{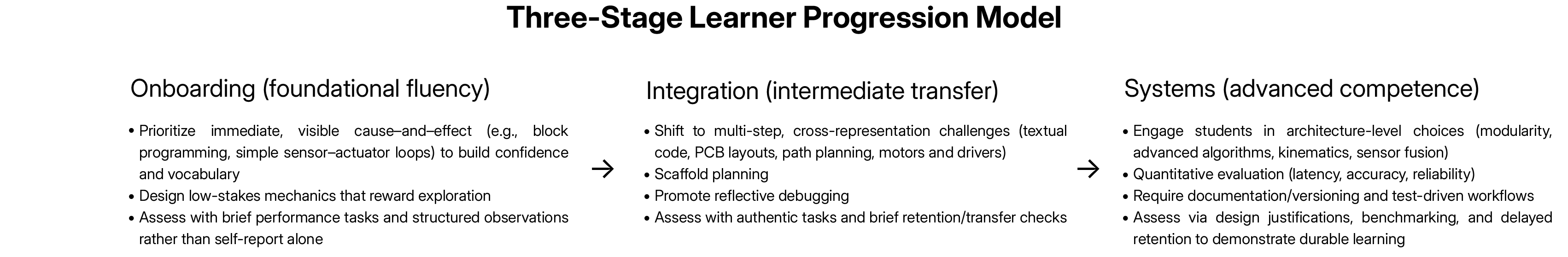}
\end{subfigure}

\caption{Design guidelines for game-mediated robotics learning. (a) Guidelines by approach and context, with quadrants listing best-bets, common pitfalls, and likely outcomes, along with representative citations. (b) Three-stage learner progression model that extends the guidelines by aligning tasks, scaffolds, and assessments with learner readiness.}
\label{fig:design_progression}\Description{A composite diagram shows a design space, comparing game-based learning and gamification across formal and informal contexts. Each section lists best practices, pitfalls, likely outcomes, and representative evidence. For formal contexts, game-based learning emphasizes early understanding but faces cost and logistics challenges, while gamification supports quantifiable skill gains with scaffolded complexity. Informal contexts highlight motivation and retention for game-based learning and inclusivity for gamification. At the bottom, a three-stage learner progression model illustrates advancement from onboarding (foundational fluency) to integration (intermediate transfer) to systems (advanced competence).}
\end{figure*}

Building on the patterns identified in this review, we introduce a structured design space to guide the development of game-mediated robotics education systems (see Figure \ref{fig:design_progression}). These guidelines outline best bets, common pitfalls, anticipated outcomes, and representative evidence for applying GBL and gamification in robotics education across both formal and informal contexts. Furthermore, drawing on our discussion in Section~\ref{discussion_skills}, these guidelines incorporate a three-staged learner progression model, enabling designers to address the continuum from novice to expert learners (see Figure \ref{fig:design_progression}). It is important to note that, as parameters such as exposure duration and instructional intensity were reported inconsistently across studies, our guidelines should be interpreted as qualitative, design-oriented patterns rather than prescriptive recommendations. Future longitudinal work is needed to derive more concrete parameter guidance.

Additionally, we identify several key directions for future research:



\begin{itemize}
    \item Develop and test adaptive, progression-aware systems that support learners’ growth from novice to expert. While many existing interventions successfully engage beginners, few systematically examine how game-mediated platforms can support sustained skill development, systems thinking, and professional identity formation over time (see Sections~\ref{results_studyChars_experience} and~\ref{discussion_skills}).
    \item Create and validate frameworks for accessible GBL and gamification in robotics education, evaluate their impact on learning and motivation across diverse populations, and identify best practices for inclusive design. As observed, relatively few studies in the current corpus explicitly addressed learners with disabilities, gender diversity, or cultural differences (see Section~\ref{discussion_equity}).
    \item Explore immersive multimodal environments, focusing on integrating haptic feedback with virtual and augmented reality systems. Although preliminary findings suggest that tactile immersion may enhance procedural learning and embodiment, rigorous comparative trials with sufficiently large and diverse samples are needed to determine the pedagogical value and cognitive-load implications of haptic-enhanced robotics learning experiences (see Sections~\ref{results_outcomes_immersive},~\ref{results_challenges_immersive}, and~\ref{discussion_modalities}).
    \item Conduct comparative studies on hybrid, multimodal learning trajectories that combine simulators, physical robots, and narrative-driven games to balance cost, accessibility, and authenticity. While hybrid trajectories are increasingly common, few studies rigorously compare their cognitive, motivational, and procedural outcomes against single-modality approaches (see Sections~\ref{results_studyChars_intervention},~\ref{results_outcomes_intervention},~\ref{results_challenges_intervention}, and~\ref{discussion_modalities}).
    \item Incorporate behavioral analytics, performance evaluations, and delayed post-tests to assess deep learning and transfer. The prevalent reliance on self-reported motivation and usability scores highlights a critical need for richer, multidimensional assessment regimes (see Section~\ref{results_challenges}).
    \item Investigate student co-design projects involving firmware, pipelines, and hardware (e.g., custom sensor boards for ROS) to enhance systems thinking, retention, and professional transfer. There was a persistent separation between advanced software instruction and hands-on robotic systems practice across the reviewed corpus (see Sections~\ref{results_studyChars_skills} and~\ref{discussion_skills}).
    \item Examine the pedagogical impact of cutting-edge AI tools, such as large language models and generative AI, in robotics education. Very few studies have incorporated these tools into game-mediated robotics learning platforms (see Section~\ref{discussion_equity}). Investigating how these technologies influence learning outcomes, student motivation, and skill development will be essential to advancing the field.
    \item Research scalable, resource-light game-mediated robotics interventions for under-resourced educational environments. While some promising examples exist (see Section~\ref{discussion_equity}), broader testing across varied socioeconomic and infrastructural contexts would enhance the generalizability of findings and inform strategies for democratizing access to robotics education globally.
\end{itemize}

Advancing the field will require methodologically rigorous, inclusivity - driven, and ecologically valid research that moves beyond short-term engagement metrics toward the sustained development of comprehensive, equitable, and scalable learning ecosystems. In addition, achieving the full potential of game-mediated robotics education will require close collaboration between designers and educators, with deliberate attention to pedagogical fit, context sensitivity, inclusivity, progression scaffolding, and evaluation rigor. Future work must move beyond proof-of-concept designs toward scalable, accessible, and pedagogically integrated learning ecosystems.


\section{LIMITATIONS}






The studies included in this review exhibited several methodological limitations that constrain the strength and generalizability of their findings. Both MMAT appraisal and ROBINS-I assessment showed that a substantial proportion of the literature lacked rigorous research designs, often omitting control groups or randomization. These shortcomings make it difficult to draw strong causal inferences regarding the effectiveness of GBL and gamified robotics education. In addition, the predominance of serious risk (particularly due to uncontrolled confounding and participant self-selection) suggests caution in attributing observed learning gains solely to the intervention effects. To address these issues, future research should adopt more robust designs, including randomized controlled trials and integrated mixed-methods approaches, and incorporate baseline equivalence checks, control groups, and validated outcome measures to reduce bias and strengthen the evidence base.

Limitations also stem from the review process itself. The search was restricted to English-language terms and four databases, potentially excluding relevant non-English or non-indexed studies. Subjective interpretation during article classification further constrain the strength of conclusions. Despite these challenges, the review offers a valuable foundation for future research, emphasizing the need for standardized reporting, diverse sampling, and multi-method evidence synthesis in game-mediated robotics education.

\section{CONCLUSIONS}

This review offers the first comparative synthesis of GBL and gamification in robotics education across 95 studies (2014–2025), coded with very high inter-rater reliability, to clarify how approach, context, pedagogy, skill emphasis, and modality jointly shape outcomes. Three robust patterns emerge: a persistent approach–context coupling (GBL more evenly distributed and common in informal settings; gamification dominant in formal classrooms), an over-emphasis on introductory programming and modular kits with comparatively little advanced software/hardware or immersive technology, and short study horizons that rely heavily on self-report. Building on these findings, we contribute a structured design space and staged progression model—linking onboarding, integration, and systems-level competence—to align mechanics, scaffolds, and assessment with learner readiness, while mapping how physical robots, simulators, digital games, and hybrid pathways afford different kinds of learning.

Taken together, the analysis reframes GBL and gamification not as competing choices but as complementary tools that can be transposed across contexts when pedagogy is treated as the mediating lever. We translate the synthesis into actionable guidance and eight concrete research directions that call for progression-aware, accessibility-first designs; deeper integration of VR/haptics where appropriate; fuller coverage of advanced software/hardware skills; and stronger methodology—triangulated measures, performance analytics, and delayed post-tests—to evidence durable gains. By coupling clear design rules with a comparative lens on outcomes, this work equips educators and designers to build inclusive, scalable, and rigorous robotics-learning experiences and provides a foundation for future studies to move beyond one-off engagements toward sustained, systems-level learning.

\section{Supplementary Materials}

Supplementary materials are available at: \href{https://osf.io/gq3a8}{https://osf.io/gq3a8/}

\begin{acks}
This research was supported in part by the National Science Foundation under Award Nos. IIS-2338122 and IIS-2548701. During the preparation of this work, the authors used Copilot and ChatGPT to check for grammar errors and improve their academic writing language. After using this tool/service, the authors reviewed and edited the content as needed and take full responsibility for the content of the publication.
\end{acks}

\graphicspath{{Icons/}}

\newcommand{\approachiconheight}{1.05em}
\newcommand{\progiconheight}{1.05em}
\newcommand{\roboticonheight}{1.05em}
\newcommand{\contexticonheight}{1.05em}
\newcommand{\uxiconheight}{1.05em}
\newcommand{\pediconheight}{1.05em}
\newcommand{\progicongap}{\hspace{0.5em}}

\newcommand{\iconraise}{-0.20em}

\newcommand{\iconsep}{\hspace{0.25em}}
\newcommand{\iconwrap}[1]{\centering\raisebox{\iconraise}{#1}}

\newcommand{\Gicon}{\iconwrap{\includegraphics[height=\approachiconheight]{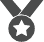}}}  
\newcommand{\GBicon}{\iconwrap{\includegraphics[height=\approachiconheight]{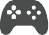}}} 

\newcommand{\PRNicon}{\iconwrap{\includegraphics[height=\progiconheight]{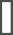}}}
\newcommand{\PRBasicon}{\iconwrap{\includegraphics[height=\progiconheight]{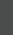}}}
\newcommand{\PRInticon}{\iconwrap{\includegraphics[height=\progiconheight]{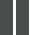}}}
\newcommand{\PRAdvicon}{\iconwrap{\includegraphics[height=\progiconheight]{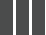}}}

\newcommand{\RNicon}{\iconwrap{\includegraphics[height=\roboticonheight]{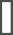}}}
\newcommand{\RBasicon}{\iconwrap{\includegraphics[height=\roboticonheight]{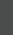}}}
\newcommand{\RInticon}{\iconwrap{\includegraphics[height=\roboticonheight]{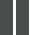}}}
\newcommand{\RAdvicon}{\iconwrap{\includegraphics[height=\roboticonheight]{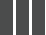}}}

\newcommand{\LIFicon}{\iconwrap{\includegraphics[height=\contexticonheight]{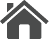}}} 
\newcommand{\LFicon}{\iconwrap{\includegraphics[height=\contexticonheight]{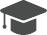}}}  

\newcommand{\UNovicon}{\iconwrap{\includegraphics[height=\uxiconheight]{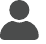}}}
\newcommand{\UExpicon}{\iconwrap{\includegraphics[height=\uxiconheight]{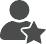}}}

\newcommand{\PedCicon}{\iconwrap{\includegraphics[height=\pediconheight]{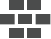}}} 
\newcommand{\PedPicon}{\iconwrap{\includegraphics[height=\pediconheight]{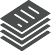}}} 
\newcommand{\PedEicon}{\iconwrap{\includegraphics[height=\pediconheight]{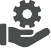}}} 

\renewcommand{\PedPicon}{\texttt{PBL}}
\renewcommand{\PedCicon}{\texttt{CON}}
\renewcommand{\PedEicon}{\texttt{EXP}}


\clearpage
\onecolumn




\begingroup

\begin{table}[htbp]
\centering
\footnotesize
\caption{Analyzed articles concerning the implementation of game-based and gamified approaches to teaching robotics and associated skills (detailed table available in the \href{https://osf.io/gq3a8/files/qzk3y}{Supplementary Materials}). 
  Approach: \GBicon{} = game-based, \Gicon{} = gamified; 
  Programming/software development skill: \PRBasicon{} = basic, \PRInticon{} = intermediate, \PRAdvicon{} = advanced, \PRNicon{} = none; 
  Robotics systems skill: \RBasicon{} = basic, \RInticon{} = intermediate, \RAdvicon{} = advanced, \RNicon{} = none; 
  Learning context: \LFicon{} = formal, \LIFicon{} = informal; 
  User experience: \UNovicon{} = novice, \UExpicon{} = experienced; 
  Pedagogical model: \PedPicon{} = project-based, \PedCicon{} = constructivist, \PedEicon{} = experiential}
\label{tab:summary}

\renewcommand{\arraystretch}{1.4}
\begin{tabular}{c c c c c c c}
\hline
\textbf{References}                      & \textbf{Approach} & \textbf{Programming/Software development skill} & \textbf{Robotics systems skill} & \textbf{Learning context} & \textbf{User experience} & \textbf{Pedagogical model} \\
\hline
\cite{eliza_game-d_2025}                 & \Gicon            & \PRInticon                                      & \RInticon                       & \LFicon                   & \UExpicon                & \PedCicon \\
\cite{budiman_gamification_2025}         & \Gicon            & \PRInticon                                      & \RInticon                       & \LFicon                   & \UExpicon                & \PedCicon \\
\cite{praveena_effective_2024}           & \Gicon            & \PRBasicon                                      & \RInticon                       & \LFicon                   & \UExpicon                & \PedEicon \\
\cite{sochol_exploration_2024}           & \GBicon           & \PRNicon                                        & \RInticon                       & \LFicon                   & \UNovicon\iconsep\UExpicon & \PedPicon\iconsep\PedEicon \\
\cite{asut_developing_2024}              & \GBicon           & \PRAdvicon                                      & \RNicon                         & \LFicon                   & \UExpicon                & \PedCicon\iconsep\PedEicon \\
\cite{bas_enhancing_2023}                & \Gicon            & \PRBasicon                                      & \RBasicon                       & \LIFicon                  & \UNovicon                & \PedPicon\iconsep\PedCicon \\
\cite{mubarrat_geobotsvr_2024}           & \GBicon           & \PRInticon                                      & \RInticon                       & \LIFicon                  & \UNovicon                & \PedEicon \\
\cite{ono_measuring_2024}                & \GBicon           & \PRBasicon                                      & \RNicon                         & \LFicon                   & \UExpicon                & \PedCicon \\
\cite{vazquez-hurtado_virtual_2024}      & \Gicon            & \PRAdvicon                                      & \RNicon                         & \LFicon                   & \UExpicon                & \PedPicon \\
\cite{osborne_wip_2024}                  & \Gicon            & \PRAdvicon                                      & \RNicon                         & \LFicon                   & \UExpicon                & \PedEicon \\
\cite{roldan-alvarez_unibotics_2024}     & \Gicon            & \PRInticon                                      & \RNicon                         & \LFicon                   & \UExpicon                & \PedPicon\iconsep\PedEicon \\
\cite{maurelli_blackpearl_2023}          & \GBicon           & \PRInticon                                      & \RInticon                       & \LFicon                   & \UExpicon                & \PedPicon\iconsep\PedEicon \\
\cite{vathanakulkachorn_enhancing_2023}  & \GBicon           & \PRBasicon                                      & \RNicon                         & \LFicon                   & \UNovicon                & \PedCicon\iconsep\PedEicon \\
\cite{stoffova_how_2023}                 & \GBicon           & \PRBasicon                                      & \RBasicon                       & \LFicon                   & \UNovicon                & \PedCicon \\
\cite{dahal_introductory_2023}           & \GBicon           & \PRBasicon                                      & \RBasicon                       & \LIFicon                  & \UNovicon                & \PedPicon\iconsep\PedCicon\iconsep\PedEicon \\
\cite{baltes_learning_2023}              & \GBicon           & \PRInticon                                      & \RInticon                       & \LIFicon                  & \UExpicon                & \PedPicon\iconsep\PedEicon \\
\cite{teogalbo_mixed_2023}               & \Gicon            & \PRAdvicon                                      & \RNicon                         & \LFicon                   & \UExpicon                & \PedPicon \\
\cite{vassigh_performance-driven_2023}   & \GBicon           & \PRAdvicon                                      & \RAdvicon                       & \LFicon                   & \UExpicon                & \PedCicon \\
\cite{de_araujo_prorobot_2023}           & \Gicon            & \PRBasicon                                      & \RBasicon                       & \LFicon                   & \UNovicon                & \PedPicon\iconsep\PedCicon\iconsep\PedEicon \\
\cite{lucan_simulator-based_2023}        & \Gicon            & \PRInticon                                      & \RNicon                         & \LFicon                   & \UExpicon                & \PedPicon\iconsep\PedEicon \\
\cite{jaggle_effectiveness_2023}         & \Gicon            & \PRBasicon\progicongap\PRInticon                    & \RBasicon\iconsep\RInticon      & \LFicon                   & \UNovicon\iconsep\UExpicon & \PedPicon\iconsep\PedEicon \\
\cite{orozco_gomez_ubuingenio_2023}      & \GBicon           & \PRBasicon                                      & \RInticon                       & \LIFicon                  & \UNovicon\iconsep\UExpicon & \PedPicon \\
\cite{vazquez-hurtado_adapting_2022}     & \GBicon           & \PRInticon                                      & \RNicon                         & \LIFicon                  & \UExpicon                & \PedPicon \\
\cite{buckley_interdisciplinary_2022}    & \GBicon           & \PRAdvicon                                      & \RNicon                         & \LFicon                   & \UExpicon                & \PedPicon\iconsep\PedEicon \\
\cite{lin_applying_2022}                 & \Gicon            & \PRInticon                                      & \RInticon                       & \LFicon                   & \UNovicon                & \PedPicon \\
\cite{fernandez-ruiz_automatic_2022}     & \Gicon            & \PRInticon                                      & \RNicon                         & \LFicon                   & \UExpicon                & \PedPicon\iconsep\PedEicon \\
\cite{peterson_evaluating_2022}          & \GBicon           & \PRAdvicon                                      & \RAdvicon                       & \LFicon                   & \UExpicon                & \PedCicon \\
\cite{kilic_exploring_2022}              & \Gicon            & \PRBasicon                                      & \RBasicon                       & \LFicon                   & \UExpicon                & \PedCicon \\
\cite{quintero_gamification_2022}        & \Gicon            & \PRBasicon                                      & \RBasicon                       & \LFicon                   & \UNovicon                & \PedPicon\iconsep\PedCicon \\
\cite{leung_project_2022}                & \Gicon            & \PRInticon                                      & \RInticon                       & \LFicon                   & \UNovicon\iconsep\UExpicon & \PedPicon \\
\cite{sarkar_teaching_2022}              & \Gicon            & \PRInticon                                      & \RNicon                         & \LFicon                   & \UNovicon\iconsep\UExpicon & \PedPicon \\
\cite{ates_work_2022}                    & \Gicon            & \PRInticon                                      & \RInticon                       & \LFicon                   & \UExpicon                & \PedEicon \\
\cite{panskyi_holistic_2021}             & \GBicon           & \PRBasicon                                      & \RInticon                       & \LIFicon                  & \UNovicon                & \PedPicon\iconsep\PedCicon \\
\cite{gharib_novel_2021}                 & \GBicon           & \PRBasicon                                      & \RBasicon                       & \LFicon                   & \UNovicon\iconsep\UExpicon & \PedPicon \\
\cite{kunovic_comparison_2021}           & \GBicon           & \PRBasicon                                      & \RBasicon                       & \LIFicon                  & \UExpicon                & \PedEicon \\
\cite{chung_design_2021}                 & \GBicon           & \PRInticon                                      & \RInticon                       & \LFicon                   & \UNovicon\iconsep\UExpicon & \PedPicon \\
\cite{yuen_mobile_2021}                  & \GBicon           & \PRBasicon                                      & \RNicon                         & \LIFicon                  & \UNovicon                & \PedPicon \\
\cite{jadhav_pbl_2021}                   & \Gicon            & \PRInticon                                      & \RInticon                       & \LFicon                   & \UNovicon\iconsep\UExpicon & \PedPicon \\
\cite{merkouris_programming_2021}        & \GBicon           & \PRBasicon                                      & \RNicon                         & \LIFicon                  & \UNovicon\iconsep\UExpicon & \PedCicon \\
\cite{nascimento_sbotics-gamified_2021}  & \Gicon            & \PRBasicon\progicongap\PRInticon\progicongap\PRAdvicon  & \RBasicon\iconsep\RInticon      & \LIFicon                  & \UNovicon\iconsep\UExpicon & \PedPicon\iconsep\PedEicon \\
\cite{peterson_teaching_2021}            & \GBicon           & \PRAdvicon                                      & \RAdvicon                       & \LFicon                   & \UExpicon                & \PedCicon \\
\cite{higashi_design_2021}               & \GBicon           & \PRBasicon                                      & \RBasicon                       & \LIFicon                  & \UNovicon                & \PedPicon\iconsep\PedCicon \\
\cite{thanyaphongphat_game-based_2020}   & \GBicon           & \PRBasicon                                      & \RBasicon                       & \LFicon                   & \UNovicon                & \PedPicon\iconsep\PedCicon\iconsep\PedEicon \\
\cite{canas_ros-based_2020}              & \Gicon            & \PRInticon                                      & \RNicon                         & \LFicon                   & \UNovicon                & \PedPicon\iconsep\PedEicon \\
\cite{panwar_analyzing_2020}             & \Gicon            & \PRInticon                                      & \RNicon                         & \LFicon                   & \UNovicon\iconsep\UExpicon & \PedPicon \\
\cite{mendonca_digital_2020}             & \GBicon           & \PRBasicon\progicongap\PRInticon                    & \RNicon                         & \LFicon                   & \UExpicon                & \PedEicon \\
\cite{giang_exploring_2020}              & \GBicon           & \PRBasicon                                      & \RBasicon                       & \LIFicon                  & \UNovicon\iconsep\UExpicon & \PedCicon \\
\cite{kashinath_narvekar_learn_2020}     & \Gicon            & \PRInticon                                      & \RInticon                       & \LFicon                   & \UNovicon\iconsep\UExpicon & \PedPicon \\
\hline
\end{tabular}
  {\captionsetup{justification=raggedleft, singlelinecheck=false}%
  \caption*{\footnotesize Continued on next page}}
\end{table}

\setcounter{table}{6}

\begin{table}[htbp]
\centering
\footnotesize
\caption[]{Analyzed articles concerning the implementation of game-based and gamified approaches to teaching robotics and associated skills (detailed table available in the \href{https://osf.io/gq3a8/files/qzk3y}{Supplementary Materials}). 
  Approach: \GBicon{} = game-based, \Gicon{} = gamified; 
  Programming/software development skill: \PRBasicon{} = basic, \PRInticon{} = intermediate, \PRAdvicon{} = advanced, \PRNicon{} = none; 
  Robotics systems skill: \RBasicon{} = basic, \RInticon{} = intermediate, \RAdvicon{} = advanced, \RNicon{} = none; 
  Learning context: \LFicon{} = formal, \LIFicon{} = informal; 
  User experience: \UNovicon{} = novice, \UExpicon{} = experienced; 
  Pedagogical model: \PedPicon{} = project-based, \PedCicon{} = constructivist, \PedEicon{} = experiential (Continued)}

\renewcommand{\arraystretch}{1.4}
\begin{tabular}{c c c c c c c}
\hline
\textbf{References}                      & \textbf{Approach} & \textbf{Programming/Software development skill} & \textbf{Robotics systems skill} & \textbf{Learning context} & \textbf{User experience} & \textbf{Pedagogical model} \\
\hline
\cite{nordin_mobile_2020}                & \GBicon           & \PRBasicon                                      & \RNicon                         & \LIFicon                  & \UExpicon                & \PedEicon \\
\cite{canas_open-source_2020}            & \Gicon            & \PRInticon                                      & \RNicon                         & \LFicon                   & \UExpicon                & \PedPicon\iconsep\PedEicon \\
\cite{wu_rapidly_2020}                   & \GBicon           & \PRInticon                                      & \RBasicon\iconsep\RInticon      & \LFicon                   & \UNovicon\iconsep\UExpicon & \PedPicon \\
\cite{sarkar_teaching_2020}              & \Gicon            & \PRInticon                                      & \RInticon                       & \LFicon                   & \UNovicon\iconsep\UExpicon & \PedPicon \\
\cite{yett2020hands}              & \GBicon            & \PRBasicon                                      & \RNicon                       & \LIFicon                   & \UExpicon & \PedPicon\iconsep\PedEicon \\
\cite{wang_codingprogramming_2019}       & \GBicon           & \PRBasicon                                      & \RBasicon                       & \LFicon                   & \UNovicon                & \PedPicon\iconsep\PedEicon \\
\cite{sharma_coding_2019}                & \GBicon           & \PRBasicon                                      & \RBasicon                       & \LIFicon                  & \UNovicon                & \PedEicon \\
\cite{diaz-lauzurica_computational_2019} & \Gicon            & \PRBasicon                                      & \RBasicon                       & \LFicon                   & \UNovicon                & \PedPicon\iconsep\PedCicon\iconsep\PedEicon \\
\cite{karageorgiou_escape_2019}          & \GBicon           & \PRBasicon                                      & \RBasicon                       & \LIFicon                  & \UNovicon\iconsep\UExpicon & \PedPicon\iconsep\PedCicon\iconsep\PedEicon \\
\cite{heljakka_gamified_2019}            & \Gicon            & \PRBasicon                                      & \RNicon                         & \LIFicon                  & \UNovicon                & \PedEicon \\
\cite{chou_interactive_2019}             & \GBicon           & \PRBasicon                                      & \RBasicon                       & \LIFicon                  & \UNovicon                & \PedCicon\iconsep\PedEicon \\
\cite{lee_learning_2019}                 & \GBicon           & \PRBasicon                                      & \RNicon                         & \LIFicon                  & \UExpicon                & \PedCicon \\
\cite{plaza_scratch_2019}                & \Gicon            & \PRBasicon                                      & \RBasicon                       & \LFicon                   & \UExpicon                & \PedPicon \\
\cite{plaza_stem_2019}                   & \Gicon            & \PRBasicon                                      & \RBasicon                       & \LFicon                   & \UExpicon                & \PedPicon \\
\cite{martin_open_2018}                  & \Gicon            & \PRInticon                                      & \RInticon                       & \LFicon                   & \UExpicon                & \PedPicon\iconsep\PedEicon \\
\cite{zainal_primary_2018}               & \GBicon           & \PRBasicon                                      & \RBasicon                       & \LIFicon                  & \UNovicon                & \PedEicon \\
\cite{matsumoto_relationship_2018}       & \GBicon           & \PRBasicon                                      & \RNicon                         & \LIFicon                  & \UNovicon                & \PedCicon \\
\cite{agalbato_robo_2018}                & \GBicon           & \PRBasicon                                      & \RNicon                         & \LIFicon                  & \UNovicon                & \PedCicon \\
\cite{hrbacek_step_2018}                 & \GBicon           & \PRBasicon\progicongap\PRInticon                    & \RInticon                       & \LFicon                   & \UNovicon                & \PedPicon\iconsep\PedEicon \\
\cite{moreno-vera_comparison_2018}       & \GBicon           & \PRBasicon                                      & \RInticon                       & \LFicon                   & \UNovicon                & \PedCicon \\
\cite{kobayashi_design_2018}             & \GBicon           & \PRInticon                                      & \RInticon                       & \LFicon                   & \UNovicon                & \PedPicon \\
\cite{hamann_gamification_2018}          & \Gicon            & \PRAdvicon                                      & \RAdvicon                       & \LFicon                   & \UNovicon                & \PedPicon\iconsep\PedEicon \\
\cite{watanabe_compulsory_2018}          & \GBicon           & \PRInticon                                      & \RInticon                       & \LFicon                   & \UNovicon\iconsep\UExpicon & \PedPicon \\
\cite{narahara2018personalizing}          & \GBicon           & \PRInticon                                      & \RBasicon                       & \LIFicon                   & \UNovicon & \PedCicon\iconsep\PedEicon \\
\cite{atmatzidou_didactical_2017}        & \Gicon            & \PRBasicon                                      & \RBasicon                       & \LFicon                   & \UNovicon                & \PedPicon\iconsep\PedCicon \\
\cite{gabriele_educational_2017}         & \GBicon           & \PRAdvicon                                      & \RNicon                         & \LFicon                   & \UNovicon                & \PedPicon\iconsep\PedCicon \\
\cite{strnad_programming_2017}           & \GBicon           & \PRInticon                                      & \RInticon                       & \LFicon                   & \UNovicon                & \PedPicon\iconsep\PedCicon\iconsep\PedEicon \\
\cite{law_teaching_2017}                 & \GBicon           & \PRBasicon                                      & \RNicon                         & \LIFicon                  & \UExpicon                & \PedCicon \\
\cite{huang_design_2017}                 & \GBicon           & \PRBasicon                                      & \RBasicon                       & \LFicon                   & \UNovicon\iconsep\UExpicon & \PedEicon \\
\cite{shim_effects_2016}                 & \GBicon           & \PRBasicon                                      & \RNicon                         & \LIFicon                  & \UNovicon                & \PedCicon\iconsep\PedEicon \\
\cite{estrada2017practical}                 & \GBicon           & \PRInticon                                      & \RBasicon                         & \LFicon                  & \UNovicon                & \PedPicon\iconsep\PedEicon \\
\cite{schmidt_computational_2016}        & \GBicon           & \PRBasicon                                      & \RNicon                         & \LIFicon                  & \UNovicon                & \PedPicon\iconsep\PedCicon \\
\cite{costa_robotics_2016}               & \GBicon           & \PRAdvicon                                      & \RNicon                         & \LIFicon                  & \UExpicon                & \PedEicon \\
\cite{han_skynetz_2016}                  & \Gicon            & \PRInticon                                      & \RNicon                         & \LIFicon                  & \UNovicon                & \PedEicon \\
\cite{hempe_erobotics_2015}              & \Gicon            & \PRAdvicon                                      & \RAdvicon                       & \LIFicon                  & \UExpicon                & \PedEicon \\
\cite{leonard_using_2016}                & \GBicon           & \PRBasicon                                      & \RBasicon                       & \LFicon                   & \UNovicon                & \PedPicon \\
\cite{yadagiri_blocks-based_2015}        & \GBicon           & \PRBasicon                                      & \RNicon                         & \LIFicon                  & \UNovicon                & \PedEicon \\
\cite{alfieri_case_2015}                 & \GBicon           & \PRBasicon                                      & \RNicon                         & \LIFicon                  & \UNovicon                & \PedCicon\iconsep\PedEicon \\
\cite{yamazaki_comparative_2015}         & \GBicon           & \PRBasicon                                      & \RNicon                         & \LIFicon                  & \UNovicon                & \PedCicon \\
\cite{duraes_gaming_2015}                & \GBicon           & \PRBasicon\progicongap\PRInticon                    & \RBasicon                       & \LFicon                   & \UNovicon                & \PedCicon \\
\cite{crespo_virtual_2015}               & \GBicon           & \PRAdvicon                                      & \RNicon                         & \LFicon                   & \UExpicon                & \PedEicon \\
\cite{crespo_virtual_2015-1}             & \GBicon           & \PRAdvicon                                      & \RNicon                         & \LFicon                   & \UExpicon                & \PedEicon \\
\cite{park_engaging_2013}                & \GBicon           & \PRBasicon                                      & \RNicon                         & \LIFicon                  & \UNovicon                & \PedPicon\iconsep\PedEicon \\
\cite{sovic_how_2014}                    & \GBicon           & \PRBasicon                                      & \RBasicon                       & \LIFicon                  & \UNovicon                & \PedCicon\iconsep\PedEicon \\
\cite{jansen_pump_2016}                  & \GBicon           & \PRBasicon                                      & \RNicon                         & \LIFicon                  & \UNovicon\iconsep\UExpicon & \PedPicon\iconsep\PedCicon\iconsep\PedEicon \\
\cite{liu_stem_2014}                     & \GBicon           & \PRInticon                                      & \RNicon                         & \LIFicon                  & \UNovicon                & \PedPicon\iconsep\PedEicon \\
\cite{roscoe_teaching_2014}              & \GBicon           & \PRBasicon                                      & \RBasicon                       & \LIFicon                  & \UNovicon                & \PedEicon \\
\hline
\end{tabular}
\end{table}

\twocolumn
\clearpage

\endgroup

\bibliographystyle{ACM-Reference-Format}
\bibliography{main}








\end{document}